\documentclass[letterpaper,journal]{IEEEtran}

\usepackage{amsmath,amsfonts,amssymb}
\usepackage{algorithm}
\usepackage[noEnd=true,indLines=true]{algpseudocodex}
\usepackage{array}
\usepackage{booktabs}
\usepackage{cite}
\usepackage{colortbl,xcolor}
\usepackage{graphicx}
\usepackage{multirow}
\usepackage[caption=false,font=normalsize,labelfont=sf,textfont=sf]{subfig}
\usepackage{stfloats}
\usepackage{tabularx}
\usepackage{textcomp}
\usepackage{url}
\usepackage{hyperref}
\usepackage{orcidlink}

\definecolor{bestcyan}{HTML}{D6F5F5}
\definecolor{secondviolet}{HTML}{EBD6F5}
\definecolor{failedgray}{HTML}{EEEEEE}
\newif\iffailrowshade
\newcommand{\failcellshade}{\iffailrowshade\cellcolor{failedgray}\fi}
\newcolumntype{F}{>{\failcellshade}c}
\newcolumntype{M}{>{\failcellshade}l}
\newcolumntype{N}{>{\failcellshade}c<{\global\failrowshadefalse}}

\newcommand{\bestavg}[1]{\multicolumn{1}{>{\columncolor{bestcyan}[0pt][0pt]}c}{#1}}
\newcommand{\secondavg}[1]{\multicolumn{1}{>{\columncolor{secondviolet}[0pt][0pt]}c}{#1}}
\newcommand{\scenename}[2]{\shortstack{#1\\#2}}

\newcommand{\failrow}{\global\failrowshadetrue}
\begin{document}

\title{TASG-Explore: Traversability-Aware Sector-Guided Exploration for Mobile Robot on Uneven Terrain}

\author{Shaocong Wang$^{\orcidlink{0009-0004-6823-2083}}$,  Shiliang Shao$^{\orcidlink{0000-0002-4512-167X}}$, Ting Wang$^{\orcidlink{0000-0001-9553-5232}}$,  Guangjie Han$^{\orcidlink{0000-0002-6921-7369}}$,~\IEEEmembership{Fellow,~IEEE}, \\
	Lianqing Liu$^{\orcidlink{0000-0002-2271-5870}}$,~\IEEEmembership{Senior Member,~IEEE}
	\thanks{Shaocong Wang is with the State Key Laboratory of Robotics, Shenyang Institute of Automation, Chinese Academy of Sciences, Shenyang 110016, China, and also with the University of Chinese Academy of Sciences, Beijing 101408, China (e-mail: wangshaocong@sia.cn).}
	\thanks{ Shiliang Shao, Ting Wang, and Lianqing Liu are with the State Key Laboratory of Robotics, Shenyang Institute of Automation, Chinese Academy of Sciences, Shenyang 110016, China.}
	\thanks{Guangjie Han is with the Key Laboratory of Maritime Intelligent Network Information Technology, Ministry of Education, Hohai University, Nanjing 210098, China.}%
}

\markboth{}%
{TASG-Explore: Traversability-Aware Sector-Guided Exploration}

\maketitle

\begin{abstract}
	Autonomous exploration on uneven terrain requires mobile robot to balance exploration efficiency, coverage completeness, and terrain safety. Detailed tsrrain reasoning improves local reliability but can slow large-scale exploration, whereas coarse region guidance expands quickly in open areas but can miss narrow passages and irregular traversable boundaries. To address this challenge, this paper presents TASG-Explore, a traversability-aware sector-guided exploration framework for mobile robot. The framework first performs hierarchical traversability analysis using variable-voxel ground fitting and adaptive 8-bit obstacle encoding. It then splitting cost map into sectors, incrementally updates sector clusters, extracts terrain-coupled frontier viewpoints, and maintains a dynamic topological roadmap with unknown topological hypotheses. Finally, a sector-guided planner selects region targets and inserts local viewpoints to generate efficient exploration routes. Benchmark experiments in diverse challenging environments, including caves, forests, and rugged hills, show that TASG-Explore achieves the best overall performance among six representative state-of-the-art planners. The proposed traversability analysis improves processing efficiency by 6.3 times while maintaining high accuracy, and the exploration planner improves exploration efficiency by 51\% and increases coverage by up to 2.95 times in rugged hill scene. Large-scale real-world experiments further demonstrate the practical value of the proposed method.
\end{abstract}


\section{Introduction}
\IEEEPARstart{G}{round} robots are widely used for autonomous inspection, search and rescue, underground exploration, and outdoor mapping. In these tasks, prior maps are often incomplete. Manual mapping can be expensive, slow, or unsafe. Autonomous exploration enables a robot to build a map while deciding where to move and what to observe next. Early methods expanded the known space by following local frontiers \cite{1,2,3,4,5}. However, local frontier utility alone does not capture the long-term travel cost induced by a sequence of exploration decisions. Recent methods increasingly reason about the global distribution of unknown space \cite{6,7,8,9}. This progress highlights a central tradeoff between local information gain and global route efficiency: a viewpoint that is locally informative may induce a long or redundant global exploration route.

Many recent advances in fast 3-D exploration have been driven by aerial-robot systems. These systems improve efficiency by coupling local viewpoint selection with global or hierarchical planning \cite{6,7,8,9,10,11,12,13,14}. However, their motion assumptions do not transfer directly to ground robots. For aerial robots, motion cost is largely determined by collision-free geometric distance, flight time, and dynamic feasibility. For ground robots, the same geometric displacement can have very different costs because reachability depends on the supporting terrain.

A common simplification in ground robot exploration is to use 2-D free space as a proxy for mobility. This assumption supports roadmap, topological-graph, frontier-clustering, and region-level exploration methods in flat or structured environments \cite{15,16,17,18,19,20}. However, in outdoor, underground, and field environments, this projection discards terrain information that is essential for ground motion. A short 2-D path may cross a steep slope, a step, or loose terrain, and a visible frontier may lie beyond unsafe or uncertain terrain.

Recent work has incorporated terrain information into ground robot exploration by using 3-D traversability constraints, terrain-aware roadmaps, or elevation-map-based region extraction \cite{21,22,23,24,25}. However, these methods still face a tradeoff between efficiency and completeness. Fine-grained traversability analysis improves safety but can lead to conservative exploration, whereas coarse region-level guidance enables fast expansion in open areas but may overlook narrow passages, irregular boundaries, or locally unsafe terrain. Existing methods therefore lack a unified mechanism that adaptively switches between coarse global guidance and fine local terrain validation.

To address these challenges, this paper proposes \textbf{TASG}-Explore, a \textbf{T}raversability-\textbf{A}ware \textbf{S}ector-\textbf{G}uided exploration framework for ground robots. The system first builds a hierarchical traversability analysis. Variable voxels rapidly fit large support surfaces and preserve wide traversable regions for efficient planning, while adaptive 8-bit encoding recovers local obstacles and non-traversable structures within each ground cell. This representation allows the planner to reason coarsely in open areas and refine terrain details only where local safety and completeness matter.

Based on this representation, we introduce incremental sector-based region segmentation to organize unknown space at a scale suitable for global exploration. The sector geometry preserves the direction, depth, and angular width of unexplored regions relative to the robot, enabling efficient estimation of region-level exploration utility. Unlike fixed partitions or pure frontier clusters, the incremental sector update keeps large unknown regions available for fast expansion while still allowing narrow or irregular regions to be separated and revisited when their local structure becomes observable.

\begin{figure*}[t]
	\centering
	\includegraphics[width=0.96\textwidth]{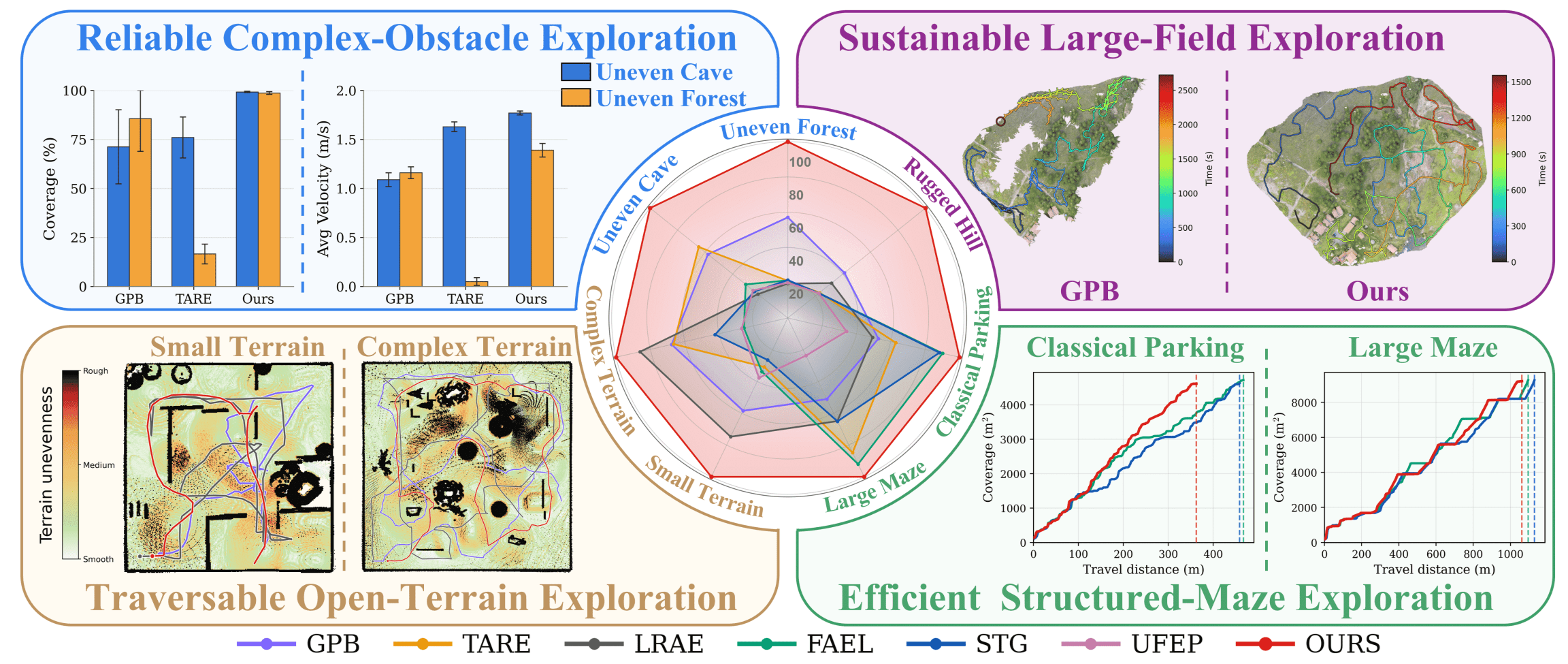}
	\caption{System performance comparison. The central radar chart summarizes exploration efficiency scores from Section~\ref{sec:simulation_benchmark} across seven scenarios, which are grouped into four categories according to the distinct challenges they pose to autonomous exploration systems: Structured-Maze, Open-Terrain, Complex-Obstacle, and Large-Field. Compared with other methods, the proposed method achieves the best overall performance in autonomous exploration tasks across different types of environments.}
	\label{fig1}
	\vspace{-0.4cm}
\end{figure*}

At the planning level, we construct a dynamic roadmap with unknown topological hypotheses. Verified traversable samples maintain safe connectivity in known space, whereas unknown samples provide temporary hypotheses toward distant candidate passages. Multi-scale sparsification compresses redundant vertices in large open regions and retains local vertices that are important for narrow structures and terrain transitions. The planner first optimizes the visiting order of sector regions to ensure efficient large-scale coverage, and then inserts frontier viewpoints to recover local observations along sector edges, narrow passages, and terrain discontinuities.

The proposed method jointly improves exploration efficiency and map coverage on uneven terrain. Open areas are explored rapidly by region-level targets and a sparse roadmap. Complex terrain and narrow spaces are handled by fine terrain validation and viewpoint insertion. To provide controlled and reproducible evaluation, we develop a comprehensive simulation-and-evaluation benchmark for uneven-terrain exploration. The benchmark includes underground caves, dense forests, and large outdoor hills. We compare the proposed method with state-of-the-art planners \cite{11,17,22,24,25,26} under the same robot, sensor, map, and termination settings. The system-level comparison in Fig.~\ref{fig1} further summarizes the exploration performance across scenarios grouped by the distinct challenges they pose to autonomous exploration systems, showing the best overall performance of the proposed method across different environment types.

The main contributions are summarized as follows:
\begin{itemize}
	\item A hierarchical traversability analysis for uneven-terrain exploration. It combines variable-voxel ground fitting with adaptive obstacle encoding, enabling the recovery of large traversable areas and fine-grained obstacle analysis.
	\item An incremental sector-based spatial segmentation. It estimates unknown-region utility from sector geometry, updates regions online, and generates terrain-coupled reachable viewpoints.
	\item A global exploration planner based on dynamic topological roadmap with unknown topological hypotheses. It maintains a multi-scale sparse roadmap and optimizes unknown sector-guided routes to support efficient large-scale exploration and complete coverage in complex scenes.
	\item A comprehensive simulation-and-evaluation benchmark for ground robot exploration on uneven terrain. Extensive simulation studies and real-world experiments validate the effectiveness of the proposed system.
\end{itemize}

\section{Related Work}
\subsection{Autonomous Exploration Methods}
Early autonomous exploration research was largely driven by frontier selection and local information gain. Yamauchi proposed frontier-based exploration, which uses the boundary between known free space and unknown space as a navigation target \cite{1}. Burgard \emph{et al.} extended frontier selection to multi-robot exploration by balancing information gain and travel cost \cite{27}. Stachniss \emph{et al.} modeled exploration actions as an information-gain maximization problem under SLAM uncertainty \cite{2}. Bircher \emph{et al.} proposed a receding-horizon next-best-view planner that enables a micro aerial vehicle to select high-gain viewpoints in local 3-D space \cite{3}. Selin \emph{et al.} and Dai \emph{et al.} improved the online efficiency of information-driven exploration for large-scale 3-D environments and MAV applications \cite{4,28}. These methods established important foundations for online exploration, but their decision making is largely local and short-horizon. The next viewpoint is usually determined by immediate information gain, visibility, and travel distance.

To mitigate short-horizon decision making, recent exploration systems explicitly maintain global context. FUEL builds an incremental frontier structure and combines it with hierarchical planning for efficient UAV exploration \cite{10}. TARE designs a hierarchical framework for complex 3-D environments by combining local exploration with a sparse global representation \cite{11}. RACER extends rapid exploration to decentralized multi-UAV systems with limited communication and local maps \cite{29}. FALCON introduces coverage-path guidance into aerial exploration and reduces repeated backtracking through route-level coverage information \cite{6}. VRExplorer organizes candidate viewpoints into viewpoint regions to reduce redundant evaluation of adjacent viewpoints \cite{7}. FLARE uses large unknown regions to guide frontier clustering and moves aerial robots toward high-potential unknown space \cite{30}. RUSH constructs a graph-based task planner for hierarchical UAV exploration in large 3-D space \cite{8}. EDEN separates coarse task reasoning from local aerial motion planning in a dual-layer framework \cite{9}.

These studies show that autonomous exploration has moved from isolated frontier selection to global route reasoning. Most of them, however, are designed for general 3-D exploration or aerial platforms. Their planning costs are usually defined by information gain, geometric distance, coverage path length, or team allocation. The terrain geometry and traversability that constrain ground motion are often omitted or treated only as downstream navigation constraints.

\subsection{Ground-Robot Exploration Methods}
Many ground robot exploration systems simplify the planning space to a 2-D occupancy grid, roadmap, or topological graph. This abstraction is natural for wheeled robots on flat ground. Once free grid cells are identified, exploration can be treated as target selection and path search on a planar graph. Wang \emph{et al.} proposed an incremental roadmap construction method that uses sampled nodes and edges to maintain long-distance reachability during exploration \cite{15}. A later semantic road map introduced semantic structure into indoor exploration, and the main planning representation remains a graph over traversable free space \cite{16}. Dang \emph{et al.} proposed a graph-based underground exploration planner that uses a persistent graph to support path planning and target selection in large subterranean spaces \cite{31}.

Recent ground robot exploration work also improves target representation and long-horizon decision quality. FAEL accelerates exploration in large-scale mobile-robot environments through efficient frontier processing and graph search \cite{17}. STGPlanner exploits global information through a skeleton topological graph and uses a finite-state machine to generate efficient coverage paths \cite{26}. Sun \emph{et al.} proposed a concave-hull-induced graph gain that is robust to frontier shape and map noise \cite{18}. CURE organizes hierarchical multi-robot exploration around unknown-region centroids \cite{19}. Region partition explicitly decomposes unknown space into regions to reduce the number of candidate targets \cite{20}. TIPS balances local and global exploration utility with a hierarchical information-rich planning strategy \cite{32}. Recent quadruped exploration work uses graph neural networks estimation to select feasible actions for legged platforms \cite{39}. These methods improve ground-robot exploration efficiency, but most reasoning is still performed on 2-D frontiers, regions, or topological graphs. This simplification works in flat or moderately structured environments. On uneven terrain, a geometrically short planar path may cross a steep slope, a step, or a discontinuous surface. A high-value 2-D frontier may correspond to an unsafe, unreachable, or terrain-incompatible viewpoint.

\subsection{Uneven-Terrain Exploration Methods}
Uneven-terrain exploration sits at the intersection of exploration planning and traversability-aware navigation. 3-D maps provide the geometric basis for this problem. Voxblox incrementally builds a Euclidean signed distance field for efficient online 3-D collision reasoning \cite{33}. UFOMap uses a probabilistic voxel map that explicitly represents unknown space, which is suitable for partially observed environments \cite{34}. For ground robots, Fankhauser \emph{et al.} proposed a robot-centric elevation map with uncertainty estimates, providing a natural representation for slope, roughness, and height-discontinuity analysis \cite{35}. These maps provide essential geometric representations, but they do not decide which unknown region should be explored next or how targets should be ordered.

Traversability-aware navigation methods further address ground motion constraints. STEP performs risk-aware off-road planning through stochastic traversability evaluation and explicitly considers uncertainty on rough terrain \cite{36}. PUTN uses plane fitting to build an uneven-terrain navigation framework for local surface reasoning \cite{37}. Agishev \emph{et al.} learned a robot-terrain interaction model and used it for trajectory optimization in large underground environments \cite{38}. These methods improve safe motion, but their primary objective is terrain navigation rather than large-scale unknown-space coverage.

Some systems connect terrain awareness with exploration target decisions. GBPlanner extends graph-based underground exploration to wheeled and legged robots, showing the importance of persistent route structures under complex terrain and limited communication \cite{22}. Azpurua \emph{et al.} proposed 3-D terrain-aware autonomous exploration for underground and confined spaces, incorporating terrain constraints into target selection \cite{21}. Zhang \emph{et al.} used aerial active exploration to find traversable paths for ground robots \cite{40}. Gao \emph{et al.} studied autonomous exploration for a bimodal aerial-ground robot in unknown confined environments \cite{41}. LRAE targets ground-robot exploration on uneven terrain and introduces large-region awareness for safe and fast planning \cite{24}. Jia \emph{et al.} proposed a portable planner to improve ground robot exploration in unknown environments \cite{42}. UFEP constructs a multi-resolution topological representation from elevation-map evidence to exploit exploration potential in complex 3-D scenes \cite{25}.

These uneven-terrain studies go beyond the flat ground assumption of conventional ground robot exploration. Traversability is no longer only a low-level navigation constraint; it becomes an important factor in exploration quality. Nevertheless, uneven-terrain exploration still faces a fundamental tradeoff between coverage efficiency and terrain fidelity. Fast large-scale exploration methods usually employ compact frontier, region, or graph representations, but such abstractions may miss narrow passages, small reachable areas, or locally complex terrain structures. Conversely, methods with detailed terrain analysis can improve safety but often lead to conservative target selection and reduced exploration efficiency. To address this tradeoff, this paper develops a compact terrain-aware exploration framework that supports efficient coverage of large unknown areas while retaining local sensitivity to terrain safety and small-scale exploratory opportunities.

\section{System Overview}
\label{sec:system_overview}
\begin{figure*}[t]
	\centering
	\includegraphics[width=0.96\textwidth]{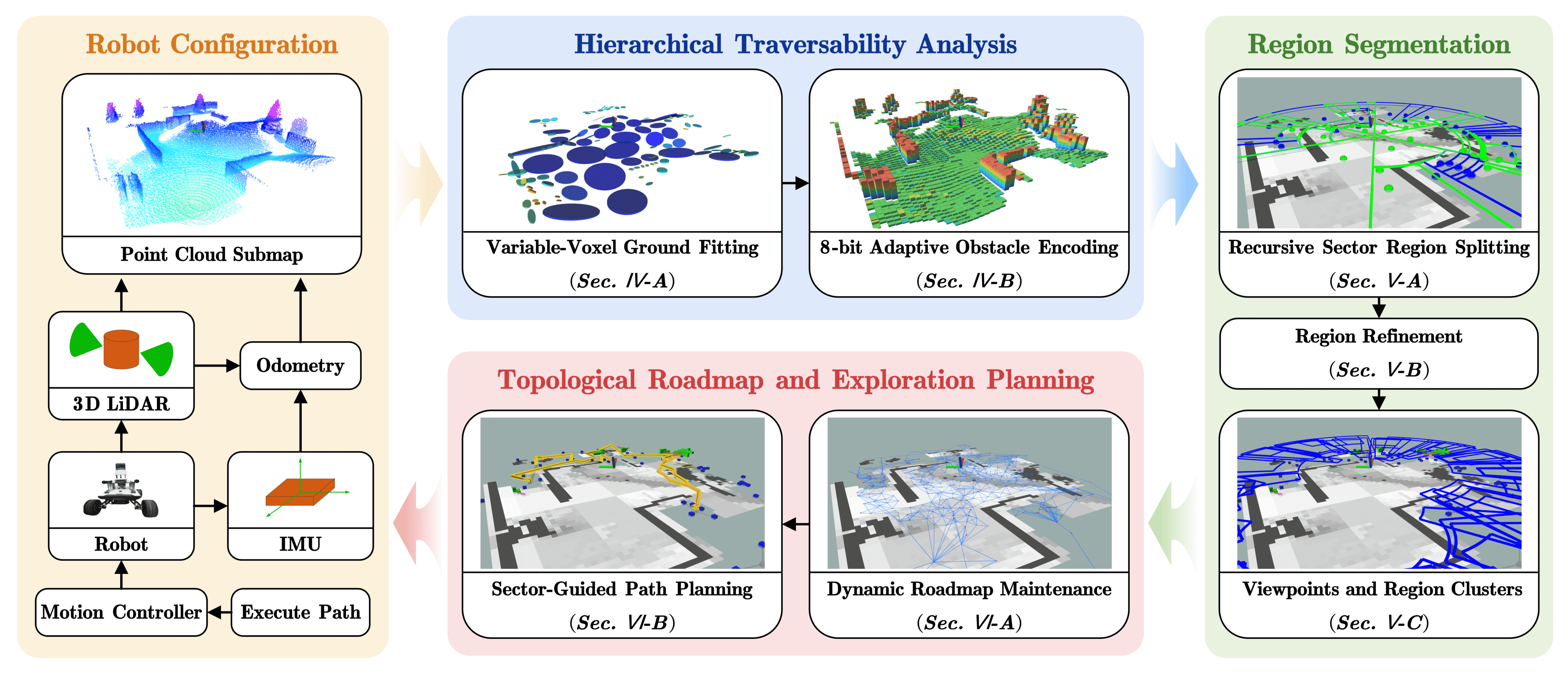}
	\caption{System overview of TASG-Explore. The system consists of hierarchical traversability analysis, incremental region segmentation, and global exploration planning. Hierarchical traversability analysis models ground surfaces and obstacles with complementary structures. Incremental sector segmentation extracts unknown regions, traversable samples, and terrain-coupled viewpoints. Global planning maintains a dynamic roadmap and generates coverage-oriented exploration routes.}
	\label{fig2}
	\vspace{-0.4cm}
\end{figure*}

\begin{table}[t]
	\centering
	\caption{NOMENCLATURE}
	\label{tab:nomenclature}
	\renewcommand{\arraystretch}{1.12}
	\begin{tabularx}{\linewidth}{@{}p{0.34\linewidth}X@{}}
		\toprule
		Symbol & Meaning \\
		\midrule
		\(\mathcal{P}_t,\mathcal{P}_{t}^{s}\) & Current LiDAR scan and local point-cloud submap. \\
		\(\mathcal{C}^{l},\mathcal{C}^{g}\) & Local and global traversability cost maps. \\
		\(\mathcal{M}^{f}\) & 3-D occupancy map for frontier and visibility queries. \\
		\(\mathcal{M}_{t}^{l}\) & Robot-centered traversability map sampled from \(\mathcal{C}^{g}\). \\
		\(\mathcal{S}(\mathbf{o},r^{-},r^{+},\theta,\Delta\theta,s)\) & Sector representation of a known or unknown local region. \\
		\(\mathcal{A}_{t}^{f},\mathcal{A}_{t}^{u},\mathcal{A}^{g}\) & Local known sectors, local unknown sectors, and retained global unknown sectors. \\
		\(P^{f},P^{u}\) & Verified traversable samples and unknown-hypothesis samples. \\
		\(V^{f},V^{u}\) & Terrain-coupled frontier viewpoints and clustered unknown-region targets. \\
		\(\mathcal{G}=(\mathcal{V}^{f}\cup\mathcal{V}^{u},\mathcal{E})\) & Dynamic roadmap with verified and hypothetical vertices. \\
		\(\sigma,\mathbf{P}_{r}\) & Region visiting order and sector-guided path before viewpoint insertion. \\
		\bottomrule
	\end{tabularx}
	\vspace{-0.4cm}
\end{table}

\subsection{Problem Definition}
This paper studies autonomous exploration for ground robots in unknown uneven environments. At each time step(\(t\))), the robot receives an odometry state \(\mathbf{x}_t=\{\mathbf{p}_t,\mathbf{R}_t\}\), where \(\mathbf{p}_t\in \mathbb{R} ^3\) is the robot position and \(\mathbf{R}_t\in SO(3)\) is the orientation. It also receives an input LiDAR point cloud \(\mathcal{P}_t\) and a local point-cloud submap \(\mathcal{P}_{t}^{s}\). The goal is to build a map online and select safe, informative future observation positions so that the robot can complete coverage in minimum exploration time.

Unlike an aerial robot, a ground robot cannot choose exploration targets based only on 3-D visibility or collision-free geometric distance. A candidate target must lie on a locally supportable terrain, and the path to the target must avoid non-traversable areas, unreliable terrain, and obstacles. The system therefore maintains two map representations. The 2.5-D global terrain cost map \(\mathcal{C}^g\) supports traversability analysis and unknown-region segmentation, where each grid cell is annotated with traversability and occupancy states. The 3-D occupied voxel map \(\mathcal{M}^f\) supports frontier extraction, with each voxel classified as occupied, unknown, or free. Table~\ref{tab:nomenclature} summarizes the main symbols used in the proposed method.

\subsection{Proposed Framework Overview}
Fig.~\ref{fig2} shows the overall framework of the proposed exploration system. The framework consists of three main modules: hierarchical traversability analysis (Section~\ref{sec:terrain_representation}), incremental region segmentation (Section~\ref{sec:region_segmentation_sampling}), and global exploration planning (Section~\ref{sec:roadmap_exploration_planning}). After receiving the latest submap from the SLAM backend, the system performs hierarchical terrain modeling in the updated region. It uses variable-resolution voxels to model large ground surfaces efficiently (Section~\ref{subsec:variable_voxel_ground_fitting}) and adaptive 8-bit obstacle encoding to represent obstacle structures above the ground (Section~\ref{subsec:height_obstacle_encoding}). These two representations are fused and incrementally written into \(\mathcal{C}^g\). The resulting traversability information supports region abstraction, connectivity queries, and path-cost evaluation (Section~\ref{subsec:terrain_cost_incremental_update}).

The region segmentation module then performs sector-based splitting around the current robot position (Section~\ref{subsec:recursive_sector_splitting}), followed by region refinement (Section~\ref{subsec:region_refinement}) and unknown-region clustering (Section~\ref{subsec:viewpoint_and_unknown_clustering}). During this process, \(\mathcal{M}^f\) is updated with LiDAR observations, and terrain-coupled frontier viewpoints are extracted. The global planning module uses the retained unknown regions, frontier viewpoints, and sample points to maintain a dynamic roadmap (Section~\ref{subsec:dynamic_roadmap_maintenance}). Finally, the sector-guided planner jointly reasons about unknown regions and frontier viewpoints, optimizes their visiting order, generates a terrain-consistent local path, and publishes the next navigation goal to the robot (Section~\ref{subsec:region_guided_exploration_planning}). The system repeats this update-and-replan process until no valid unknown regions or frontier viewpoints remain.

\section{Hierarchical Traversability Analysis}
\label{sec:terrain_representation}
\begin{figure*}[t]
	\centering
	\includegraphics[width=0.96\textwidth]{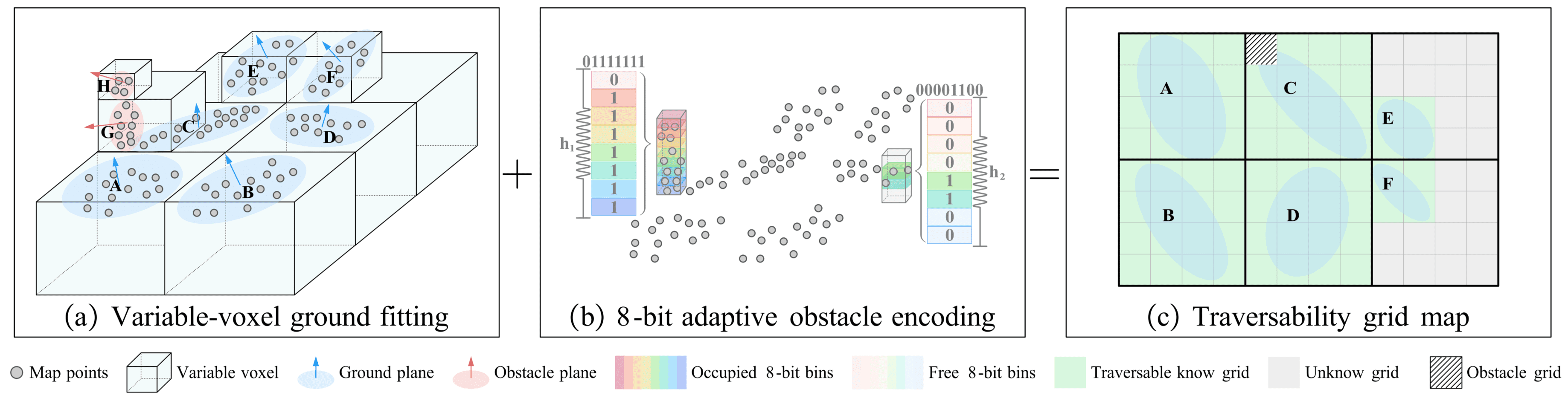}
	\caption{Hierarchical traversability analysis. (a) Variable voxels are used to fit traversable ground, enabling fast, large-scale traversability analysis while reducing unknown holes in the map. (b) An 8-bit adaptive obstacle encoding is used to perform fine-grained analysis of obstacles in the scene, ensuring precise obstacle segmentation. (c) A local traversability cost grid map is obtained by integrating the two modeling approaches.}
	\label{fig3}
	\vspace{-0.4cm}
\end{figure*}

When a new local submap around the robot is received, the global terrain cost map \(\mathcal{C}^g\) is incrementally updated by a local cost map \(\mathcal{C}^l\). As shown in Fig.~\ref{fig3}, the key idea is to use two complementary representations. Variable-resolution voxels retain occupancy evidence from the point cloud and support large-scale plane fitting. Obstacle encoding provides fine obstacle information for ground robot planning.

Both \(\mathcal{C}^g\) and \(\mathcal{C}^l\) store traversability cost \(c\), maximum point-cloud height \(z^{+}\), minimum point-cloud height \(z^{-}\), ground plane normal \(\mathbf{n}\), and plane offset \(d\). The cost value \(c=-1\) denotes unknown terrain, \(0\le c<1\) denotes known traversable terrain, and larger values indicate higher traversal cost. The value \(c=1\) denotes non-traversable terrain.

\subsection{Variable-Voxel Ground Fitting}
\label{subsec:variable_voxel_ground_fitting}
As shown in Fig.~\ref{fig3}(a), the system first fits ground planes from \(\mathcal{P}_{t}^{s}\). Following the data structure in \cite{43}, \(\mathcal{P}_{t}^{s}\) is organized as a hash-indexed variable-voxel submap \(\mathcal{M}_t^{v}\). Each voxel is initialized with resolution \(r_v\) and can be recursively split. A voxel that can be represented stably by a single plane is kept as a coarse unit. A voxel containing non-planar structure is divided into eight subvoxels until a stable plane is found or the maximum split depth \(L\) is reached. Thus, open smooth ground can be represented by large planes without unknown holes, while steps, obstacle boundaries, and rough regions are described at finer resolution.

For a voxel \(v\) containing points \(\{\mathbf{p}_{i}\}_{i=1}^{N}\), the centroid and covariance are
\begin{equation}
	\boldsymbol{\mu}_{v}=\frac{1}{N}\sum_{i=1}^{N}\mathbf{p}_{i},\qquad
	\boldsymbol{\Sigma}_{v}=
	\frac{1}{N}\sum_{i=1}^{N}
	(\mathbf{p}_{i}-\boldsymbol{\mu}_{v})
	(\mathbf{p}_{i}-\boldsymbol{\mu}_{v})^{\mathrm{T}}.
	\label{eq:voxel_covariance_zh}
\end{equation}
Principal component analysis is then applied to \(\boldsymbol{\Sigma}_{v}\). Let \(\lambda_{1}\le \lambda_{2}\le \lambda_{3}\) be the eigenvalues, and let \(\mathbf{e}_{1}\) be the eigenvector corresponding to the smallest eigenvalue. When \(\lambda_{1}<\tau_{p}\), the voxel is considered a stable plane. The plane normal and offset are \(\mathbf{n}_{v}=\mathbf{e}_{1}\) and \(d_{v}=-(\mathbf{n}_{v})^{\mathrm{T}}\boldsymbol{\mu}_{v}\), respectively. The plane covariance is obtained by propagating the point covariance in \(v\), and it represents the uncertainty of the plane normal and center. The detailed propagation follows \cite{43}.

After plane extraction, each 2-D cell in \(\mathcal{C}^l\) is initialized from its lowest local surface. The length and width of \(\mathcal{C}^l\) are determined by the extrema of \(\mathcal{P}_{t}^{s}\) along the \(x\) and \(y\) directions. For each grid cell \(\alpha\), the system queries the  hash-index of lower sample \(\mathbf{p}_{\alpha}^{-}(x_\alpha,y_\alpha,z^-_\alpha)\) in \(\mathcal{M}_t^{v}\) and retrieves its plane. If the plane exists, the traversability criteria are computed as
\begin{equation}
	\eta_{\alpha}^{-}\!=\arccos(|\mathbf{e}_{z}^{\mathrm{T}}\mathbf{n}_{\alpha}^{-}|),\quad
	u_{\alpha}^{-}\!=\mathrm{clip}_{[0,1]}\left[ \left( \frac{\mathrm{tr}(\boldsymbol{\Sigma}_{\alpha}^{-})}{\sigma_u} \right)^{\gamma_u} \right].
	\label{eq:ground_cost_zh}
\end{equation}
where, \(\eta_{\alpha}^{-}\) is the angle between the plane normal and gravity direction, and \(u_{\alpha}^{-}\) is the normalized plane roughness. The parameter \(\sigma_u\) scales roughness, and \(\gamma_u\) controls nonlinear sensitivity. If \(\eta_{\alpha}^{-}<\eta_{\max}\) and \(u_{\alpha}^{-}<u_{\max}\), the lowest plane is treated as traversable ground. The cell cost is initialized as \(c_{\alpha}= \omega\eta_{\alpha}^{-}/\eta_{\max}+(1-\omega)u_{\alpha}^{-}\), and the ground parameters are assigned from the lower plane, where $\omega=0.7$. Cells without a valid lower plane remain unknown. This initialization restores large traversable surfaces and reduces holes in the map while keeping ground fitting efficient.

\subsection{Adaptive 8-bit Obstacle Encoding}
\label{subsec:height_obstacle_encoding}
The initialized \(\mathcal{C}^l\) mainly contains large traversable cells and unknown cells. As shown in Fig.~\ref{fig3}(b), each 2-D grid cell is assigned an adaptive 8-bit height code to place obstacle evidence back onto \(\mathcal{C}^l\).

For a grid cell \(\alpha\), the encoding state is defined as
\begin{equation}
	\mathcal{H}_{\alpha}
	=
	\{z_{\alpha}^{\max}, \Delta r_{\alpha}, \mathbf{E}_{\alpha} \},
	\qquad
	\mathbf{E}_{\alpha}\in\{0,1\}^{B}.
	\label{eq:height_code_state_zh}
\end{equation}
where, \(z_{\alpha}^{\max}\) is the maximum observed height within the robot traversability height range \([ z_{\alpha}^{-}, z_{\alpha}^{-}+h_{\mathrm{trav}}]\), \(\Delta r_{\alpha}\) is the adaptive vertical resolution, and \(\mathbf{E}_{\alpha}\) is a $B$-dimensional binary occupancy code with bits ordered from lower to higher vertical layers. In this paper, $B=8$, representing one byte of encoded data. Since \(z_{\alpha}^{\max}\) is obtained by traversing \(\mathcal{P}_{t}^{s}\), the encoding range and scale update dynamically with \(z_{\alpha}^{\max}\):
\begin{equation}
	\Delta r_{\alpha}\!=\max \!\left( \frac{z_{\alpha}^{\max}-z_{\alpha}^{-}}{B},h_{\mathrm{occ}} \right),\quad
	z_{\alpha}^{o}\!=\frac{z_{\alpha}^{-}+z_{\alpha}^{\max}-B\Delta r_{\alpha}}{2}.
\end{equation}
where, \(h_{\mathrm{occ}}\) is the maximum obstacle height that the robot can pass over, and \(z_{\alpha}^{o}\) is the height corresponding to the bottom layer of \(\mathbf{E}_{\alpha}\).

To reduce submap traversal, the system uses an iterative reprojection encoding strategy. All elements in \(\mathcal{H}_{\alpha}\) are updated in one pass. When a new point increases \(z_{\alpha}^{\max}\) and changes \(\Delta r_{\alpha}\), the previous occupancy code is not cleared. Instead, it is reprojected onto the new layers according to interval overlap:
\begin{equation}
	\begin{aligned}
		E_{\alpha,k}\!&=\bigvee_{l=0}^{B-1}{\left( \check{E}_{\alpha,l}\land \mathbb{I} \!\left[ \check{I}_{\alpha,l}\cap I_{\alpha,k}\ne \emptyset \right] \right)},\\
		I_{\alpha,k}\!&=[z_{\alpha}^{o}+k\Delta r_{\alpha}, z_{\alpha}^{o}+\left( k+1 \right) \Delta r_{\alpha}).
	\end{aligned}
\end{equation}
where, \(\check{E}_{\alpha,l}\) and \(\check{I}_{\alpha,l}\) are the previous occupancy state and height interval of the \(l\)-th bit, and \(I_{\alpha,k}\) is the current height interval of the \(k\)-th bit. This reprojection preserves accumulated obstacle evidence even when the vertical scale of the grid cell changes during submap traversal.

After the encoding is constructed, \(\mathbf{E}_{\alpha}\) refines \(c_{\alpha}\) to produce the local cost map in Fig.~\ref{fig3}(c). If many occupied intervals exist between the lower surface and \(z_{\alpha}^{\max}\), the cell is directly marked as an obstacle. If only a few occupied intervals exist, the system queries the \(\mathcal{M}_t^{v}\) at the lowest occupied height and evaluates the upper plane. Near-vertical height discontinuities are treated as obstacles. Continuous but rough slopes are assigned higher cost rather than being removed from the traversable set. The refined local cost is summarized as
\begin{equation}
	c_{\alpha}=\begin{cases}
		1,&		|\mathcal{K}_{\alpha}|\ge B/2 \\
		1,&		h_{\mathrm{occ}}\leqslant \Delta h\leqslant 2h_{\mathrm{trav}}\\
		\omega c_{\alpha}+ (1-\omega)c_{\alpha}^{u},&
		\exists \Pi_{\alpha}^{u},\ \Delta\eta_{\alpha}<\eta_{\max}/2, \eta_{\alpha}^{u}<\eta_{\max}\\
		\max \left( c_{\alpha},c_{\alpha}^{u} \right),&
		\exists \Pi_{\alpha}^{u} ,\ \Delta\eta_{\alpha}\geqslant\eta_{\max}/2, \eta_{\alpha}^{u}<\eta_{\max}\\
		c_{\alpha},&		\mathrm{otherwise}.
	\end{cases}
\end{equation}
where, \(\mathcal{K}_{\alpha}=\{k\mid E_{\alpha,k}=1, k>k^-\}\) is the set of upper occupied indices in \(\mathbf{E}_{\alpha}\), and \(k^-=\max \left( \left\lceil (z_{\alpha}^{-}-z_{\alpha}^{o})/\Delta r_{\alpha}\right\rceil,0 \right)\). The value \(z_{\alpha}^{u}\) is the lowest occupied height. If \(\mathcal{K}_{\alpha}\) is empty, \(z_{\alpha}^{u}=z_{\alpha}^{\max}\); otherwise, \(z_{\alpha}^{u}=z_{\alpha}^{o}+\min(\mathcal{K}_{\alpha})\Delta r_{\alpha}\). Let \(\Pi_{\alpha}^{u}\) denote the upper plane queried at \(z_{\alpha}^{u}\), if such a plane exists. Its normal is \(\mathbf{n}_{\alpha}^{u}\), and \(\Delta\eta_{\alpha}=\arccos(|(\mathbf{n}_{\alpha}^{-})^{\mathrm{T}}\mathbf{n}_{\alpha}^{u}|)\). The height difference is \(\Delta h=\|\mathbf{p}_{\alpha}^{u}-\mathbf{p}_{\alpha}^{-}\|\), \(\eta_{\alpha}^{u}\) is the angle between the upper-plane normal and gravity, and \(c_{\alpha}^{u}=\omega\eta_{\alpha}^{u}/\eta_{\max}+(1-\omega)u_{\alpha}^{u}\) is the traversal cost of the upper plane.

\subsection{Terrain Cost Incremental Update}
\label{subsec:terrain_cost_incremental_update}
After \(\mathcal{C}^{l}\) is updated, it is incrementally fused into the fixed-size \(\mathcal{C}^{g}\). If a local grid cell \(\alpha\) projects to a global grid cell \(\beta\), the global cost is updated by
\begin{equation}
	c_{\beta}^{g}
	=
	\begin{cases}
		c_{\alpha}, & c_{\beta}^{g}=-1,\ c_{\alpha}\neq -1,\\
		(1-\omega_{\alpha})c_{\beta}^{g}+\omega_{\alpha}c_{\alpha},
		& c_{\beta}^{g}\in[0,1),\ c_{\alpha}\neq -1,\\
		c_{\beta}^{g}, & \text{otherwise},
	\end{cases}
	\label{eq:global_cost_update_zh}
\end{equation}
where \(\omega_{\alpha} =\min(\omega,10/\rho_{\beta})\) is a local fusion weight selected according to observation reliability and the distance \(\rho_{\beta}\) from the grid cell to the robot. The same projection also updates global point-cloud height, ground normal, and plane offset when the local cell is traversable. Therefore, the global terrain map supports obstacle and free-space queries, and it provides surface-pose information for terrain-coupled viewpoint generation and topological edge validation.

Dynamic objects and sparse observations can leave false obstacles in the global map. To reduce these artifacts, each global grid cell maintains an obstacle reliability \(\varphi_{\beta}\) and the number of consecutive free observations \(m_{\beta}\). They are updated as
\begin{align}
	\varphi_{\beta,t}
	&=
	\begin{cases}
		\min(\varphi_{\max}, \varphi_{\beta,t-1}+2\Delta\varphi),
		& c_{\alpha}\ge 0.9,\\
		\max(0, \varphi_{\beta,t-1}-\Delta\varphi),
		& 0\le c_{\alpha}<0.9,
	\end{cases}
	\label{eq:reliability_update_zh}\\
	m_{\beta,t}
	&=
	\begin{cases}
		0, & c_{\alpha}\ge 0.9,\\
		m_{\beta,t-1}+1, & 0\le c_{\alpha}<0.9.
	\end{cases}
	\nonumber
\end{align}
where, \(\Delta\varphi=1\) is a update step size. When a global cell marked as occupied is repeatedly observed as free and \(\varphi_{\beta}\) drops below the retention threshold, the old obstacle is removed and replaced by the new local observation. This filter keeps real walls and terrain breaks stable while allowing short-term obstacles caused by moving objects or sparse measurements to disappear.

\section{Region Segmentation and Sampling}
\label{sec:region_segmentation_sampling}
The region segmentation module converts the continuously updated terrain map into exploration information. It outputs four types of results: traversable samples in known regions, unknown-region clusters, terrain-coupled frontier viewpoints, and hypothetical samples in unknown regions. The robot maintains a sliding window of size \(W^s\times W^s\) centered at its current position. It samples the global cost map \(\mathcal{C}^{g}\) in real time to obtain a local traversability map \(\mathcal{M}_t^l\). According to traversability cost, grid cells in \(\mathcal{M}_t^l\) are labeled as Free (\(0\leqslant c_{\beta}^{g}<1\)), Occupied (\(c_{\beta}^{g}=1\)), or Unknown (\(c_{\beta}^{g}=-1\)). We represent each segmented region as a sector:
\begin{equation}
	\begin{aligned}
		\mathcal{S}(\mathbf{o},r^-,r^+,\theta,\Delta \theta,s)
		=
		\{&\,\mathbf{o}+\tau [\cos \omega ,\sin \omega ]^{\mathrm{T}}\mid\\
		&\tau \in [r^-,r^+],\,\omega \in [\theta ,\theta +\Delta \theta ]\}.
	\end{aligned}
	\label{eq:sector_def_zh}
\end{equation}
where, \(\mathbf{o}\in \mathbb{R}^2\) is the center of \(\mathcal{M}_t^l\), \(r^-\) and \(r^+\) are the inner and outer radii, and \(\theta\) and \(\Delta\theta\) are the start angle and angular width. The state \(s\in\{0,1\}\) indicates the region type. A known sector has \(s=0\) and contains Free or Occupied cells. An unknown sector has \(s=1\) and contains Unknown cells. As shown in Fig.~\ref{fig4}(a), sector regions naturally preserve observation angle \(\vartheta\), radial depth\(r^+-r^-\), and angular width \(\Delta\theta\). They are therefore suitable for describing the distribution of unknown space within a local sensor range.

\subsection{Recursive Sector Region Splitting}
\label{subsec:recursive_sector_splitting}
\begin{figure*}[t]
	\centering
	\includegraphics[width=0.96\textwidth]{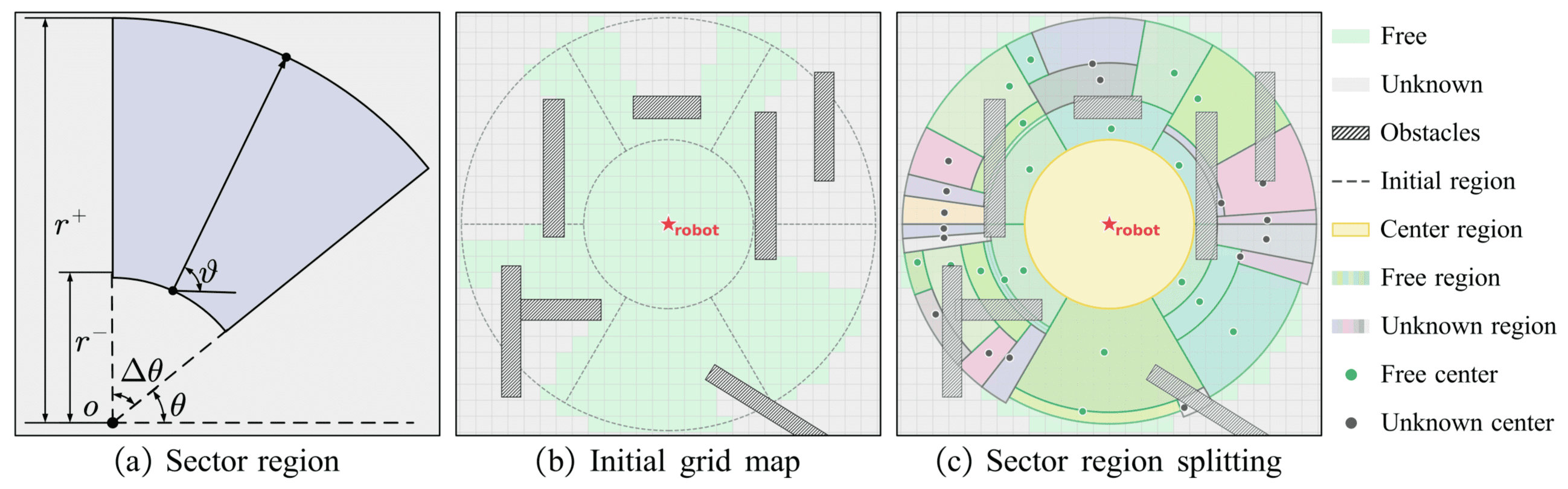}
	\caption{Sector region segmentation. (a) Elements of a sector region. (b) Initial candidate sectors. (c) Sector regions after recursive splitting.}
	\label{fig4}
	\vspace{-0.4cm}
\end{figure*}

Sector segmentation is performed by a recursive process with state flipping, as shown in Algorithm~\ref{alg:sector_segmentation}. The algorithm first searches outward from the local-map center \(\mathbf{o}\) to obtain the inner boundary \(r_{\min}\), avoiding repeated processing of covered space near the robot (lines~\ref{ln:seg_init0}--\ref{ln:seg_init1}). The local space is then initialized as six candidate sectors with angular width \(\pi/3\) and state \(s=0\), as shown in Fig.~\ref{fig4}(b). These sectors are pushed into queue \(\mathcal{Q}\) (line~\ref{ln:seg_init1}). For each candidate sector, the current state \(s\) and its complement \(\bar{s}\) are set as the positive state \(s^{+}\) and negative state \(s^{-}\). The algorithm also initializes the sample accumulator, valid sample number, and split flag (line~\ref{ln:seg_state}).

The sector is scanned in polar coordinates with radial step \(r_l\) and angular step \(1^\circ\). The algorithm counts samples in the positive state and the number of cells \(N_{\bar{s}}\) in the negative state (lines~\ref{ln:seg_accum}--\ref{ln:seg_count}). When \(N_{\bar{s}}\) exceeds an arc-length threshold, a state transition is detected and the split position \((\tau_s,\omega_s)\) is recorded (lines~\ref{ln:slip1}--\ref{ln:slip2}). If the valid sample number is \(N=0\), the initial state assumption is considered invalid. The algorithm flips the state and recomputes the samples, which improves robustness (lines~\ref{ln:n01}--\ref{ln:n02}). Otherwise, the inner region \(\tau<\tau_s\) is treated as a valid sector \(\mathcal{S}_{\mathrm{in}}\), and its representative point \(\mathbf{p}_{\mathrm{in}}\) is computed as the sample mean. The sector is inserted into either the known-sector set \(\mathcal{A}_t^{f}\) or the local unknown-sector set \(\mathcal{A}_t^{u}\) according to its final state (lines~\ref{ln:n11}--\ref{ln:n12}). If the split is detected, the outer region \(\tau\geq\tau_s\) is split into two candidate sectors that inherit the current and complementary states. They are pushed back into the queue for further processing (lines~\ref{ln:nb1}--\ref{ln:nb2}). This recursive splitting and state-flipping mechanism allows the algorithm to adapt to known/unknown boundaries in complex environments, producing stable and detailed local region segmentation.

\begin{algorithm}[t]
	\centering
	\caption{Recursive Sector Region Splitting}
	\label{alg:sector_segmentation}
	\small
	\setlength{\baselineskip}{13pt}
	\begin{algorithmic}[1]
		\Require local map $\mathcal{M}_t^{l}$
		\Ensure known-sector set $\mathcal{A}_t^{f}$, unknown-sector set $\mathcal{A}_t^{u}$
		\State $r_{o}\gets$\texttt{getInnerBoundary}($\mathcal{M}_t^{l}$,$\mathbf{o}$)\label{ln:seg_init0}
		\State $r_{\min}\gets \max(r_{o}-4r_l,0)$,$r_{\max}\gets{{\sqrt{2}W^s}/{2}}$
		\State $\mathcal{Q}\gets\{\mathcal{S}(\mathbf{o},r_{\min},r_{\max},i\pi/3,\pi/3,s=0)\}_{i=0}^{5}$ \label{ln:seg_init1}
		\State $\mathcal{A}_t^{f}\gets\emptyset$, $\mathcal{A}_t^{u}\gets\emptyset$
		\While{$\mathcal{Q}$ is not empty}
		\State Pop $\mathcal{S}=(\mathbf{o}, r^{-},r^{+},\theta,\Delta\theta,s)$ from $\mathcal{Q}$
		\State $s^{+} = s$, $s^{-} =\bar{s}$, $\mathbf{p}_s^{\Sigma}=[0,0]^{\mathrm{T}}$, $N = 0$,  $b\gets\mathrm{false}$ \label{ln:seg_state}
		\For{$\tau=r^{-}:r_{l}:r^{+}$}  \label{ln:seg_accum}
		\State $N_{\bar{s}} = 0$
		\For{$\omega=\theta:1^{\circ}:\theta+\Delta\theta$}
		\State $\mathbf{q} = \mathbf{o}+\tau [\cos \omega ,\sin \omega ]^{\mathrm{T}}$
		\State \textbf{if} $\mathcal{M}_t^{l}(\mathbf{q})\in s^{+}$ \textbf{then} $\mathbf{p}_s^{\Sigma}+=\mathbf{q}$, $N++$
		\State \textbf{if} $\mathcal{M}_t^{l}(\mathbf{q})\in s^{-}$ \textbf{then} $N_{\bar{s}}++$ \label{ln:seg_count}
		\If{$N_{\bar{s}}>\max \left( 5,\tau (\omega-\theta) /6r_{l} \right) $}\label{ln:slip1}
		\State $(\tau_{s},\omega_{s})\gets(\tau-r_{l},\omega-1^{\circ})$, $b\gets\mathrm{true}$
		\State break out of two \textbf{for} loops\label{ln:slip2}
		\EndIf
		\EndFor
		\EndFor
		\If{$N=0$}\label{ln:n01}
		\State $s=\bar{s}$; \texttt{recompute} $(\mathbf{p}_s^{\Sigma},N,b,\tau_{s},\omega_{s})$ with $s^{+}$, $s^{-}$\label{ln:n02}
		\EndIf
		\If{$N>0$}\label{ln:n11}
		\State $\mathcal{S}_{\mathrm{in}}\gets(\mathbf{o},r^{-},\tau_{s},\theta,\Delta\theta,s)$, $\mathbf{p}_{\mathrm{in}}=\mathbf{p}_s^{\Sigma}/N$
		\State $(s=0\ ?\ \mathcal{A}_t^{f}:\mathcal{A}_t^{u}).\texttt{insert}(\mathcal{S}_{\mathrm{in}},\mathbf{p}_{\mathrm{in}})$ \label{ln:n12}
		\EndIf
		\If{$b=\mathrm{true}$}\label{ln:nb1}
		\State $\mathcal{S}_{1}\gets(\mathbf{o},\tau_{s},r^{+},\theta,\omega_{s}-\theta,s)$, $\mathcal{Q}$.\texttt{insert}$(\mathcal{S}_{1})$
		\State $\mathcal{S}_{2}\gets(\mathbf{o},\tau_{s},r^{+},\omega_{s},\theta+\Delta\theta-\omega_{s},\bar{s})$, $\mathcal{Q}$.\texttt{insert}$(\mathcal{S}_{2})$\label{ln:nb2}
		\EndIf
		\EndWhile
	\end{algorithmic}
\end{algorithm}

\subsection{Region Refinement}
\label{subsec:region_refinement}
\subsubsection{Known-Region Sampling}
\label{subsubsec:known_region_sampling}
After region splitting, known regions in \(\mathcal{A}_t^{f}\) are used to generate candidate samples on traversable terrain. This process does not use uniform Cartesian grid sampling. Instead, it samples according to sector geometry in polar coordinates. The radial direction is sampled with a fixed step \(\Delta \tau=5r_l\). The angular sampling interval \(\Delta \phi\) decreases linearly with sample radius \(\tau_s\). It starts from \(\Delta\phi_\mathrm{max}\) at the inner boundary \(r^-\) and decreases to \(\Delta\phi_\mathrm{min}\) at the outer boundary \(r^+\). This design produces denser angular coverage at longer range. A sample is inserted into the known sample set \(P^f\) only if it satisfies the exploration-boundary constraint, lies in a Free local-map cell, has no nearby occupied cells, and remains within the local-map range. For a sample indexed by \(\beta\), its three-dimensional form is \(\mathbf{p}_{\beta}=[x_{\beta}^{g}, y_{\beta}^{g},z_{\beta}^{g}+h_{s}]^{\mathrm{T}}\), where \(z_{\beta}^{g}\) is the ground height of the corresponding grid cell and \(h_s\) is the sensor height.

\subsubsection{Unknown-Region Filtering}
\label{subsubsec:unknown_region_filtering}
To improve efficiency and avoid frequent global region recomputation, the global unknown-sector set \(\mathcal{A}^{g}\) is incrementally updated from \(\mathcal{A}_t^{u}\). When a new unknown sector is added to \(\mathcal{A}^{g}\), historical sectors within its \(\gamma_s\)-neighborhood are removed to limit the number of global regions. If its inner boundary is not blocked by occupied cells, the representative point is moved to the center of the inner boundary so that it stays closer to known space.

\begin{figure}[t]
	\centering
	\includegraphics[width=\linewidth]{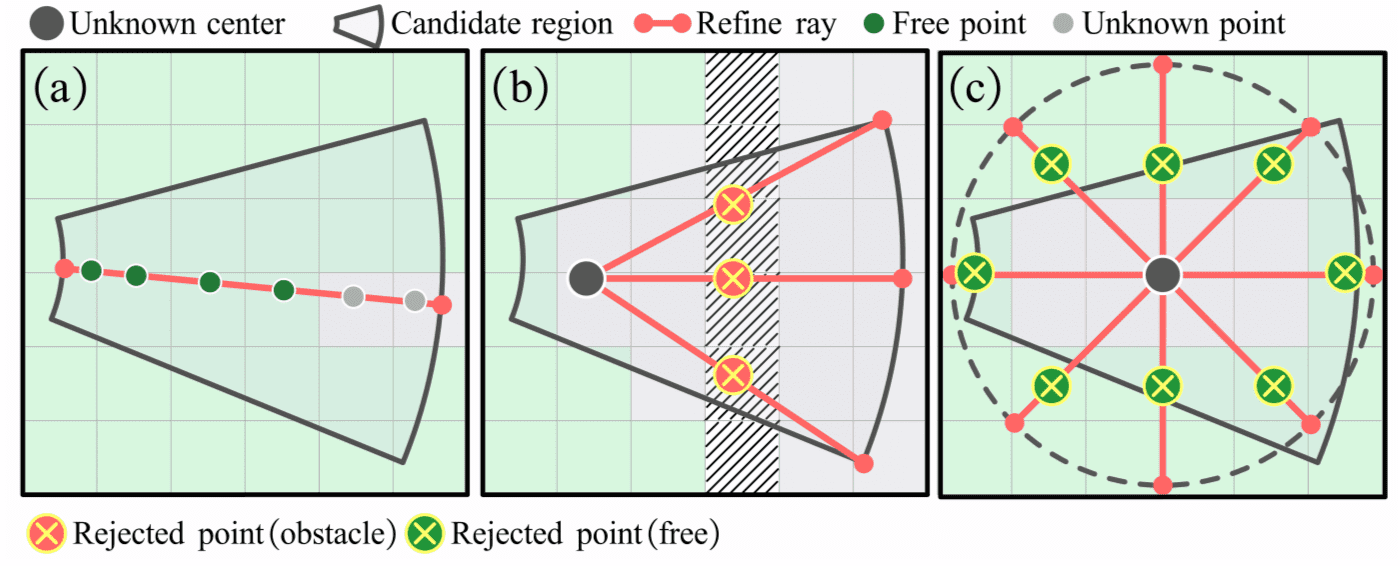}
	\caption{Ray-based filtering of unstable unknown regions. (a) A candidate region is removed when its radial unknown ratio is insufficient. (b) An unknown region behind an obstacle is removed when three outward rays are blocked. (c) An isolated hole inside known space is removed when eight surrounding rays meet known cells.}
	\label{fig5}
	\label{fig:ray_filter_zh}
	\vspace{-0.4cm}
\end{figure}

Because the distribution of unknown cells in \(\mathcal{C}^{g}\) changes online, historical unknown regions that no longer have exploration value must be filtered. As shown in Fig.~\ref{fig:ray_filter_zh}, the proposed ray-based filter handles three cases. First, if a candidate region contains too few unknown cells, it is deleted. To avoid traversing all cells in the region, the algorithm samples radial rays in the sector and removes the region when the unknown ratio along these rays is below 30\%. Second, if known cells exist near the unknown center, the algorithm casts three rays from the center to the two outer angular corners and the midpoint of the outer arc. If all three rays are blocked by occupied cells in the global terrain map, the center cannot effectively observe the area behind the obstacle and should not be retained as a target. Third, the algorithm removes isolated holes extracted from inside known regions. It casts rays from the unknown center along eight-neighborhood directions. If every ray reaches known cells within a fixed distance, the unknown region is treated as a hole surrounded by known terrain rather than an opening to unexplored space. Regions with too small a coverage radius, too narrow an angular span, or centers too close to obstacles are also removed.

\subsection{Terrain-Coupled Frontier Viewpoints and Unknown Region Clustering}
\label{subsec:viewpoint_and_unknown_clustering}
\subsubsection{Viewpoint Extraction}
\label{subsubsec:viewpoint_extraction}
Known samples in \(P^f\) are first used to extract three-dimensional frontier viewpoints for local frontiers. The algorithm updates \(P^f\) by removing samples that are too close to candidate viewpoints used in the previous planning cycle, while those previous candidate viewpoints are inserted into \(P^f\). This both supplements viewpoints in newly expanded known regions and avoids frequent viewpoint jitter across frames.

Similar to \cite{10}, the system maintains a UFOMap \(\mathcal{M}^{f}\) and incrementally updates the three-dimensional frontier set \(\mathcal{F}\). Each frontier point in \(\mathcal{F}\) searches \(P^f\) for the nearest sample within radius \(W^s\) that satisfies the constraints. A viewpoint \(\mathbf{v}\) is associated with a frontier \(\mathbf{f}\) only when visibility and reachability constraints are both satisfied. The visibility constraint requires \(\mathrm{FREE}_{\mathcal{M}^{f}}(\mathbf{v},\mathbf{f})=1\) and \(\mathrm{FREE}_{\mathcal{C}^{g}}(\mathbf{v},\mathbf{f})=1\), ensuring that the viewpoint can observe the frontier without collision. The reachability constraint checks whether the robot can move from the viewpoint toward its associated frontier over compatible terrain. Given the ground normal \(\mathbf{n}^v\) and plane offset \(d^v\) at the grid cell of the viewpoint \(\mathbf{v}\) in \(\mathcal{C}^{g}\), the relation between viewing direction and terrain is constrained by
\begin{equation}
	\eta^f
	=
	\sin^{-1}
	\frac{|(\mathbf{n}^v)^{\mathrm{T}}\mathbf{f}+d^v|}
	{\|\mathbf{f}-\mathbf{v}\|}.
	\label{eq:terrain_view_angle_zh}
\end{equation}
where, \(\eta^f\) is the angle between the line from \(\mathbf{v}\) to \(\mathbf{f}\) and the local ground plane at \(\mathbf{v}\). The frontier is considered reachable from \(\mathbf{v}\) only when \(\eta^f<\eta_{\max}\). The final frontier viewpoint set is \(V^f=\{(\mathbf{v},g^f)\}\), where \(g^f\) is the viewpoint gain defined by the number of associated frontiers.
\begin{figure}[t]
	\centering
	\includegraphics[width=\linewidth]{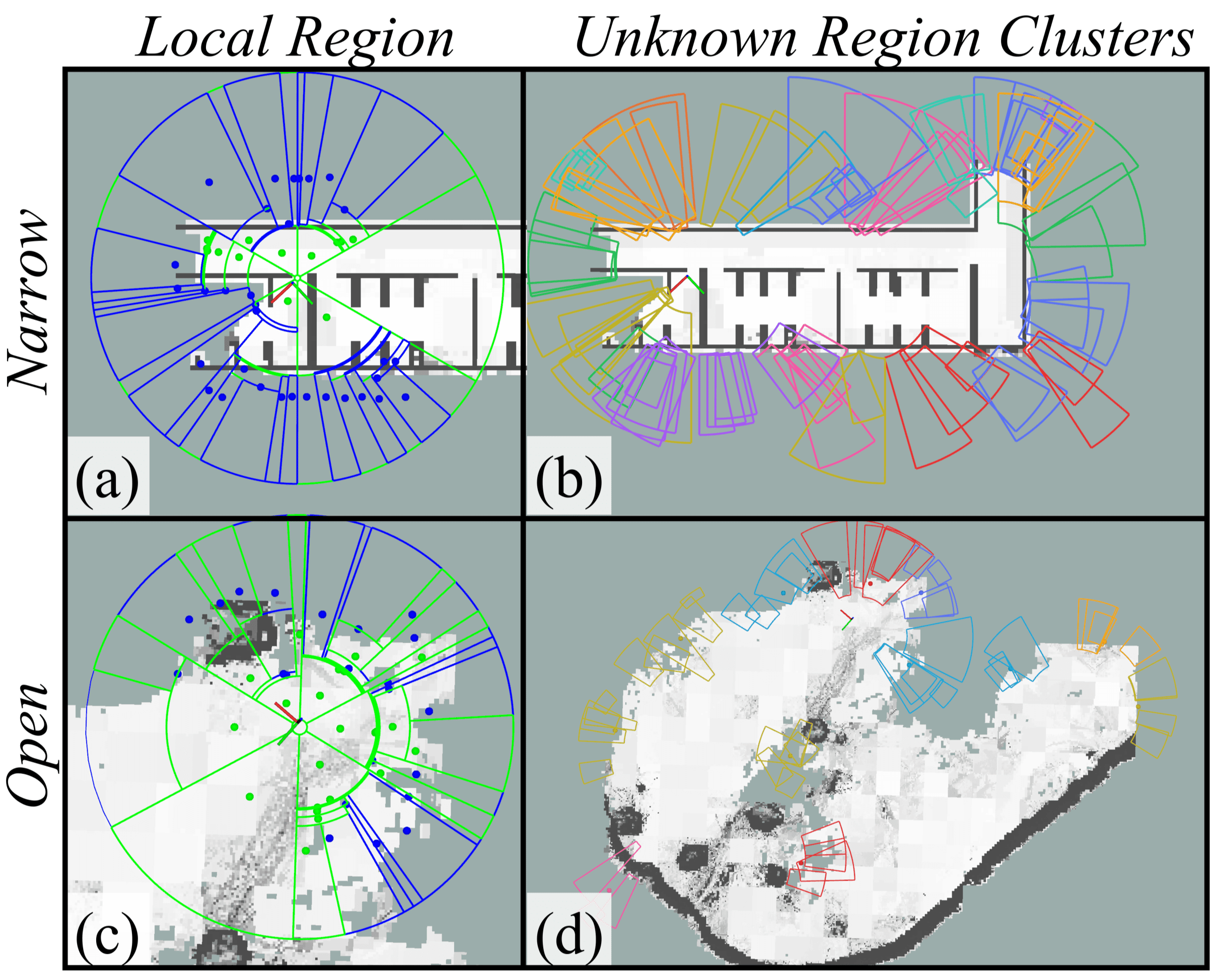}
	\caption{Sector segmentation and unknown-region clustering. (a) Sector segmentation in a narrow environment. Blue denotes unknown regions and green denotes known regions. (b) Unknown-region clustering in the narrow environment. The same color indicates the same cluster. (c) Sector segmentation in an open environment. (d) Unknown-region clustering in the open environment.}
	\label{fig:cluster}
	\vspace{-0.4cm}
\end{figure}

\begin{algorithm}[t]
	\caption{Clustering of Unknown Sectors}
	\small
	\setlength{\baselineskip}{13pt}
	\label{alg:unknown_clustering}
	\begin{algorithmic}[1]
		\Require global unknown sectors $\mathcal{A}^{g}$, previous center set
		$\mathcal{U}_{c,t-1}$
		\Ensure cluster set $\mathbb{U}_{t}$, center set
		$\mathcal{U}_{c,t}$
		\State $\mathcal{S}_{\mathrm{vis}}\gets\emptyset,\ \mathbb{U}_{t}\gets\emptyset,\ \mathcal{U}_{c,t}\gets\emptyset$
		\If{$\mathcal{U}_{c,t-1}\neq\emptyset$}\label{ln:0}
		\State $\mathcal{H}_{k}\gets\emptyset,\ k=1,\ldots,|\mathcal{U}_{c,t-1}|$
		\For{$(\mathcal{S}_{i},\mathbf{p}_{i})\in\mathcal{A}^{g}$}
		\State $\bar{\mathbf{u}}_{j,t-1}\gets$\texttt{findNearestCenter}$(\mathbf{p}_{i},\mathcal{U}_{c,t-1})$
		\If{$\|\mathbf{p}_{i}-\bar{\mathbf{u}}_{j,t-1}\|<d_{c}\land
			\mathrm{FREE}_{\mathcal{C}^{g}}(\bar{\mathbf{u}}_{j,t-1},\mathbf{p}_{i})=1$ }
		\State $\mathcal{H}_{j}\gets\mathcal{H}_{j}\cup\{\mathcal{S}_{i}\}$, $\mathcal{S}_{\mathrm{vis}}\gets\mathcal{S}_{\mathrm{vis}}\cup\{\mathcal{S}_{i}\}$
		\EndIf
		\EndFor
		\State $\mathbb{U}_{t}\gets\mathbb{U}_{t}\cup\{\mathcal{H}_{k}\mid |\mathcal{H}_{k}|>0\}$\label{ln:1}
		\EndIf
		\For{$(\mathcal{S}_{i},\mathbf{p}_{i})\in\mathcal{A}^{g}\setminus\mathcal{S}_{\mathrm{vis}}$}\label{ln:2}
		\State $(\mathcal{B},\mathcal{S}_{\mathrm{vis}})\gets\texttt{growBFSClusters}(\mathcal{S}_{i},\mathbf{p}_{i},\mathcal{A}^{g},\mathcal{S}_{\mathrm{vis}})$ \label{ln:clust_bfs}\label{ln:clust_neighbor}
		\State $\mathbb{U}_{t}\gets\mathbb{U}_{t}\cup\{\mathcal{B}\}$\label{ln:3}
		\EndFor
		\While{$\exists\,\mathcal{B}_{a},\mathcal{B}_{b}\in\mathbb{U}_{t}$}\label{ln:4}
		\If{\texttt{isMergeableCluster}$(\mathcal{B}_{a},\mathcal{B}_{b})$ }
		\State$\mathbb{U}_{t}\gets(\mathbb{U}_{t}\setminus\{\mathcal{B}_{a},\mathcal{B}_{b}\})\cup\{\mathcal{B}_{a}\cup\mathcal{B}_{b}\}$ \label{ln:clust_merge}\label{ln:5}
		\EndIf
		\EndWhile
		\For{$\mathcal{U}_{k,t}\in\mathbb{U}_{t}$}
		\State $\mathcal{U}_{c,t}$.insert(\texttt{computeClusterCenters}$(\mathcal{U}_{k,t})$) \label{ln:clust_center}
		\EndFor
		\State \textbf{return} $V^u=\{(\bar{\mathbf{u}}_{k,t}, \mathcal{U}_{k,t})\mid k=1,\ldots,|\mathcal{U}_{c,t}|\}$
	\end{algorithmic}
\end{algorithm}

\subsubsection{Unknown Region Clustering}
\label{subsubsec:unknown_region_clustering}
Filtered unknown sectors in \(\mathcal{A}^{g}\) are clustered to avoid fragmented targets. Fig.~\ref{fig:cluster} shows local region segmentation and global unknown-region clustering in two scenarios. As shown in Algorithm~\ref{alg:unknown_clustering}, clustering is incremental. Historical clusters are not discarded at every cycle. If previous cluster centers exist, they are used as priors for a fast Euclidean clustering step that preserves temporal information. Each sector \(\mathcal{S}_i\) first searches for the nearest historical center \(\bar{\mathbf{u}}_{j,t-1}\). If their distance is below the clustering threshold \(d_c\) and the line between them is obstacle-free, \(\mathcal{S}_i\) is assigned to that historical cluster, which is then inherited in the current cycle (lines~\ref{ln:0}--\ref{ln:1}). Unassigned regions are clustered by breadth-first search. Adjacent regions are grouped only when their centers are within \(d_c\) and have obstacle-free line of sight (lines~\ref{ln:2}--\ref{ln:3}). Finally, clusters are checked for merging. If two cluster centroids are mutually visible and the clusters contain at least one pair of regions whose distance is below half the threshold with no obstacle between them, the clusters are merged (lines~\ref{ln:4}--\ref{ln:5}). The resulting unknown-region clusters and their centers form \(V^u=\{(\bar{\mathbf{u}}_{k,t}, \mathcal{U}_{k,t})\}\), which is used for sector-guided planning.
\begin{figure*}[t]
	\centering
	\includegraphics[width=0.96\textwidth]{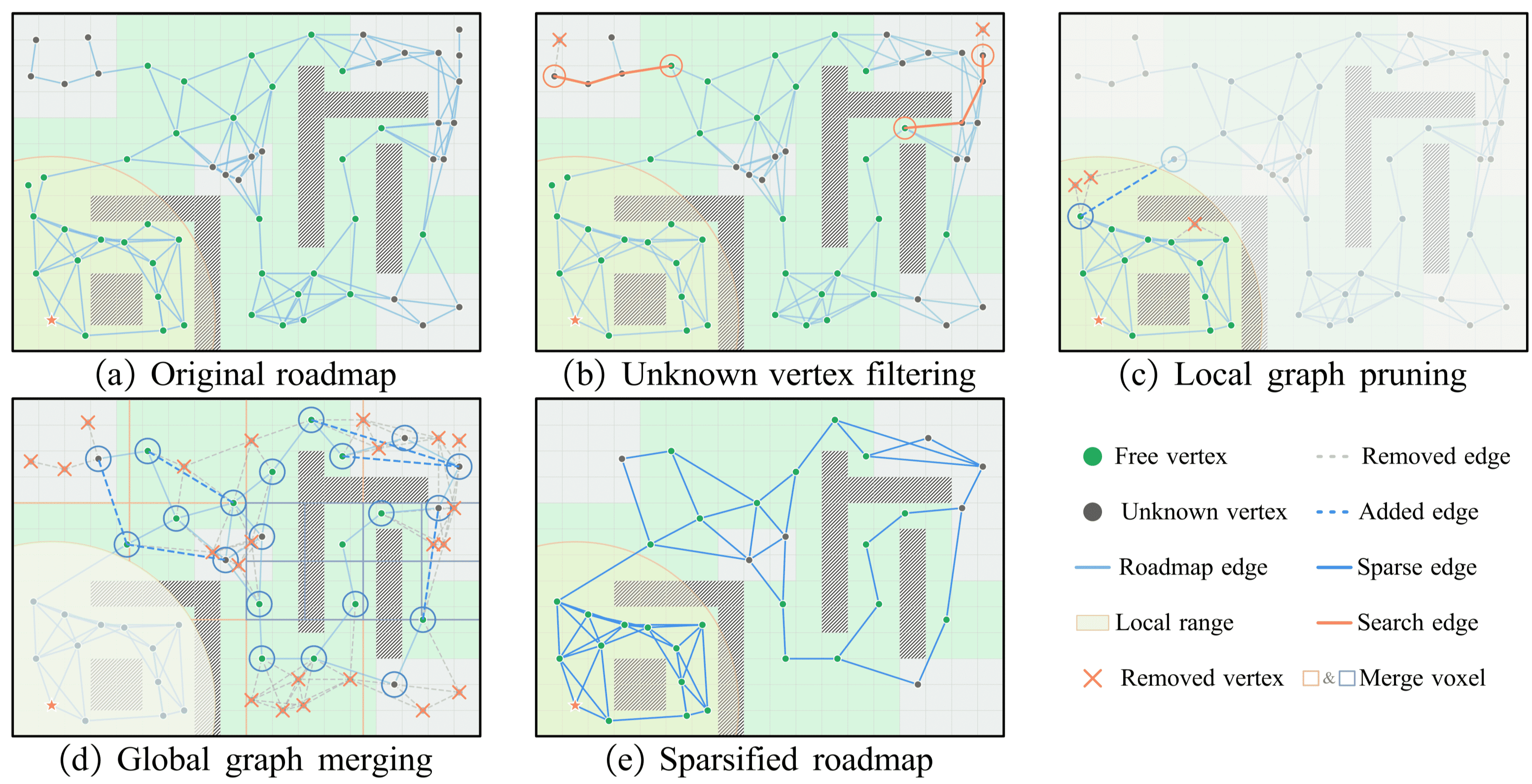}
	\caption{Dynamic topological roadmap maintenance. (a) Raw roadmap. (b) Unknown-vertex update. (c) Local pruning. (d) Global graph merging. (e) Sparse roadmap after maintenance.}
	\label{fig6}
	\label{fig:roadmap_management_zh}
	\vspace{-0.4cm}
\end{figure*}

The clustered unknown regions also generate the unknown sample set \(P^u\), which is passed to the dynamic topological roadmap. This set uses the same three-dimensional point form as the known sample set \(P^f\) in Section~\ref{subsubsec:known_region_sampling}. Its height is initialized from the nearest known grid cell. However, the two sets have different meanings. Known samples are verified traversable vertices, while unknown samples are topological hypotheses leading to unexplored space. Each unknown cluster contributes one cluster-center sample and several region-center samples. For an unknown cluster \(\mathcal{U}_{k,t}\), if the number of internal unknown regions is below five, all region centers are used. If the cluster contains more than five regions, farthest-point sampling keeps five region centers to avoid redundant unknown vertices in the roadmap.

\section{Topological Roadmap and Exploration Planning}
\label{sec:roadmap_exploration_planning}

\subsection{Dynamic Roadmap Maintenance}
\label{subsec:dynamic_roadmap_maintenance}
The planner uses a dynamically maintained roadmap to construct topological paths efficiently. The roadmap is represented as an undirected graph \(\mathcal{G}=(\mathcal{V}^{f}\cup\mathcal{V}^{u},\mathcal{E})\). It maintains two types of vertices. The set \(\mathcal{V}^{f}\) is updated from \(P^f\) and reconstructs the topology of explored traversable space. The set \(\mathcal{V}^{u}\) is updated from \(P^u\) and provides temporary hypotheses about unknown space, allowing the robot to estimate connectivity between known and unknown regions during exploration. Similar to \cite{15}, a new sample is inserted only when it is sufficiently far from existing local graph vertices. It connects to at most \(k_c\) collision-free neighbors within radius \(r_c\). After new vertices and edges are created, the graph is updated by three operations, as shown in Fig.~\ref{fig:roadmap_management_zh}. These operations keep the roadmap lightweight while preserving the main environmental structure.

\begin{figure*}[t]
	\centering
	\includegraphics[width=0.96\textwidth]{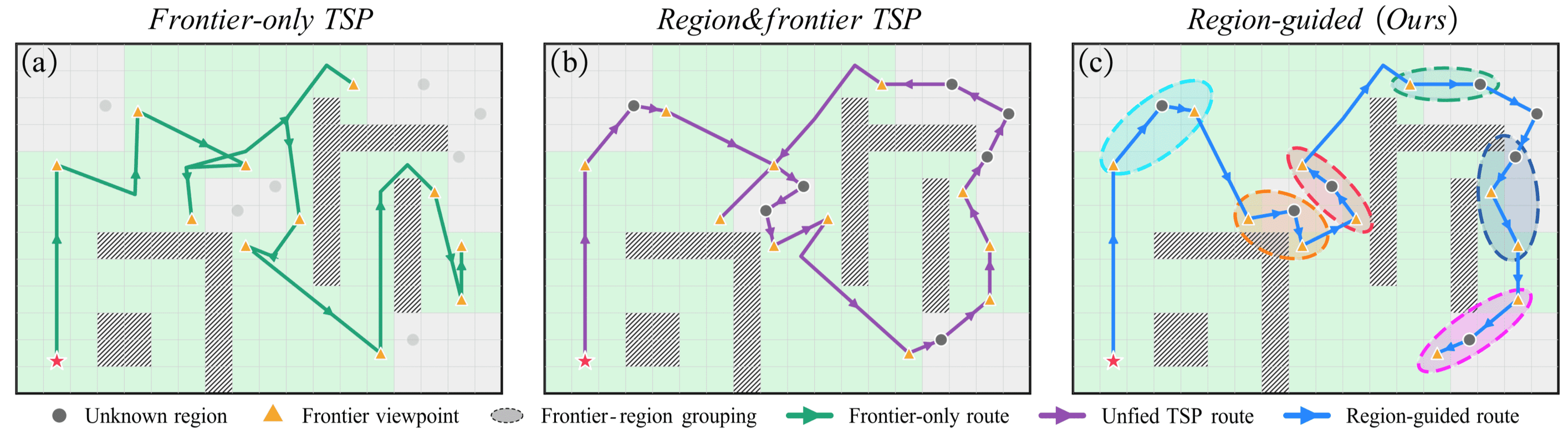}
	\caption{Comparison of exploration-route organization. (a) Viewpoint-only TSP treats all frontier viewpoints as independent targets. (b) Mixed region-viewpoint TSP optimizes targets from two scales at the same level. (c) The proposed method first orders unknown regions and then inserts associated viewpoints within each region.}
	\label{fig7}
	\vspace{-0.4cm}
\end{figure*}
\subsubsection{Unknown Vertex Updating}
\label{subsubsec:unknown_vertex_updating}
Each unknown vertex \(v_i^u\in\mathcal{V}^{u}\) first undergoes a state update. If its cost in \(\mathcal{C}^{g}\) becomes Free, it is transferred to \(\mathcal{V}^{f}\). If its cost becomes Occupied, \(v_i^u\) and its incident edges are removed from \(\mathcal{G}\). Any edge crossing an obstacle is also removed. The remaining unknown vertices then undergo a reachability update. A depth-limited search is performed on \(\mathcal{G}\). If \(v_i^u\) can reach a vertex in \(\mathcal{V}^{f}\) within three topological steps, it is considered well supported by known space. Otherwise, it is removed. As shown in Fig.~\ref{fig:roadmap_management_zh}(b), this mechanism removes isolated unknown vertices far from the exploration boundary while preserving useful connectivity hypotheses between known and unknown regions.

\subsubsection{Local Graph Pruning}
\label{subsubsec:local_graph_pruning}
The roadmap follows a near-dense, far-sparse principle. Far regions can be represented by a sparse roadmap that provides approximate global guidance. Regions near the robot keep a denser roadmap so that the robot can safely reach narrow spaces. Therefore, the local region within radius \(r_{\max}\) around the robot is not sparsified globally. Only redundant vertices and their edges in \(\mathcal{V}^{u}\) and \(\mathcal{V}^{f}\) are pruned. Vertices in the local window are handled according to graph degree. Isolated vertices and leaf vertices are deleted. A degree-two relay vertex is removed when its two neighbors can be directly connected by a bypass edge, and the bypass edge is added. Junction vertices with degree at least three are retained. This operation compresses redundant chains into single traversable edges without destroying key branches or bridges in the local roadmap, as illustrated in Fig.~\ref{fig:roadmap_management_zh}(c).

\subsubsection{Global Graph Merging}
\label{subsubsec:global_graph_merging}
Remote vertices outside the local window are sparsified so that path planning remains fast as the explored area grows. As shown in Fig.~\ref{fig:roadmap_management_zh}(d), the two vertex sets \(\mathcal{V}^{f}\) and \(\mathcal{V}^{u}\) are downsampled separately using variable-resolution voxels, thereby preventing unknown topological hypotheses from being merged into representatives of known free space. The process starts with coarse voxels of side length \(3r_c\), quickly merging redundant vertices in open remote regions. Within each voxel and each vertex type, the vertex with the highest degree is kept as the representative. External edges connected to deleted vertices are transferred to the representative. If a vertex cannot be merged because the line to the representative is blocked, it is placed into a pending set. The voxel is then split into finer voxels if the maximum split level has not been reached. The side length at level \(\ell\) is \(3r_c/2^\ell\), and each voxel can split at most \(L_m\) times. This process continues until all mergeable vertices are absorbed or the maximum split level is reached. Open areas are compressed aggressively, while obstacle boundaries and complex passages retain finer graph resolution. After preliminary merging, a breadth-first search checks the connected components of retained vertices. If a deleted vertex connects two retained vertices, that bridge vertex is restored. The final sparse roadmap in Fig.~\ref{fig:roadmap_management_zh}(e) contains far fewer vertices while preserving the major topology needed for global path queries.

\subsection{Sector-Guided Exploration Planning}
\label{subsec:region_guided_exploration_planning}
\subsubsection{Guided Path Generation}
\label{subsubsec:guided_path_generation}
The planner first constructs a sector-guided path from the unknown cluster set \(V^u=\{(\bar{\mathbf{u}}_{k,t},\mathcal{U}_{k,t})\}\). The path points are the cluster centers of the unknown regions. For an unknown-region center \(\bar{\mathbf{u}}_{i,t}\), its exploration gain is
\begin{equation}
	g_{i}^{u}
	=
	\frac{\max_{\mathcal{S}_{j}\in\mathcal{U}_{i,t}}
		(r^{+}_{j}-r^{-}_{j})}
	{\sqrt{2}W^s/2}
	\sum_{\mathcal{S}_{j}\in\mathcal{U}_{i,t}}
	r^{+}_{j}\Delta\theta_j .
	\label{eq:area_gain_zh}
\end{equation}
This gain considers both radial depth and angular breadth. Since unknown regions mainly guide paths toward broad coverage, angular breadth is computed as the sum of outer arc lengths in the cluster. This better reflects the maximum potential coverage. Radial depth is represented by the maximum sector depth and scales the breadth gain, encouraging the planner to consider distant unknown space rather than only wide nearby regions.

The path between two unknown-region centers is obtained by A* search on the roadmap. For a path \(\mathcal{P}_{ij}=(\mathbf{p}_{0},\ldots,\mathbf{p}_{L_{ij}})\), the traversal cost is
\begin{equation}
	c_{ij}
	=
	\sum_{k=0}^{L_{ij}-1}
	\psi_{k,k+1}\left(
	1+\frac{ \mathcal{C}^{g}(\mathbf{p}_{k})+ \mathcal{C}^{g}(\mathbf{p}_{k+1})}{2}
	\right)
	\|\mathbf{p}_{k+1}-\mathbf{p}_{k}\|,
	\label{eq:terrain_path_cost_zh}
\end{equation}
\begin{equation}
	\psi_{k,k+1}
	=
	\omega+(1-\omega)\log_{2}(1+\Delta\bar{\vartheta}_{k,k+1}/\pi),
	\label{eq:terrain_path_cost_zh2}
\end{equation}
where, the traversal cost of an unknown cell is assigned from the nearest known cell. The term \(\psi_{k,k+1}\) is a heading-consistency coefficient, and \(\Delta\bar{\vartheta}_{k,k+1}\) is the difference between the observation angles of two unknown regions. This coefficient encourages smoother guide trajectories with fewer yaw changes.

Let the region visiting order be \(\sigma=(\sigma_0,\sigma_1,\ldots,\sigma_{N_u})\), where \(\sigma_0\) denotes the current robot position. The final TSP objective is
\begin{equation}
	J(\sigma)=
	\sum_{m=1}^{N_u}
	g_{\sigma_m}^{u}
	\exp\left(
	-3\sum_{r=1}^{m}c_{\sigma_{r-1},\sigma_r}
	\right).
	\label{eq:discounted_utility_zh}
\end{equation}
The optimization produces a sector-guided path \(\mathbf{P}_r=(\mathbf{p}_0,\mathbf{p}_1,\ldots,\mathbf{p}_{N_u})\), where \(\mathbf{p}_0\) is the current robot position.

\begin{figure*}[t]
	\centering
	\includegraphics[width=0.96\textwidth]{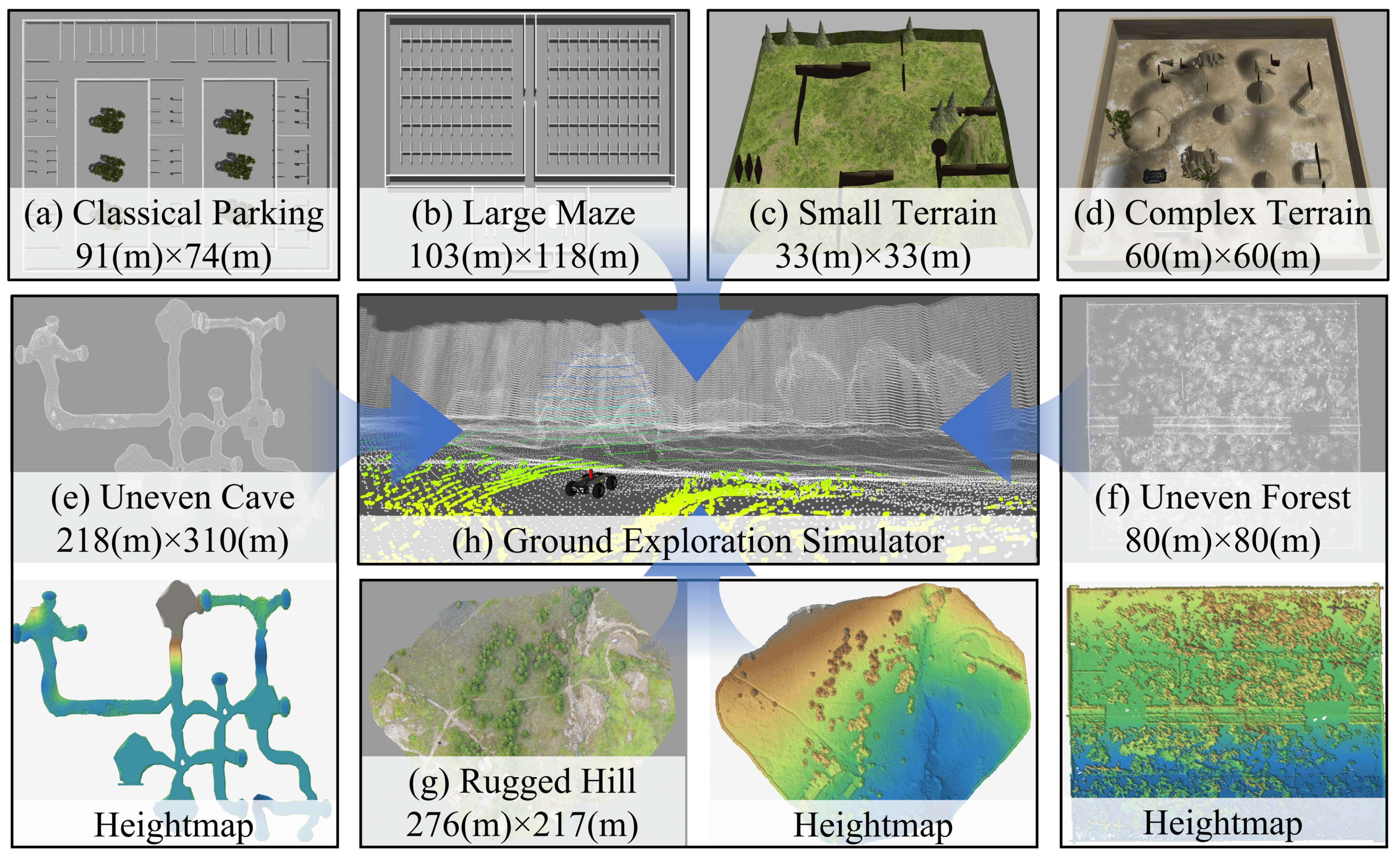}
	\caption{Benchmark environments. (a)--(f) show representative test maps, including Classical Parking, Large Maze, Small Terrain, Complex Terrain, Uneven Cave, Uneven Forest, and Rugged Hill. (g) The developed exploration simulation platform, which uses point-cloud maps as simulation environments.}
	\label{fig:robo}
	\vspace{-0.4cm}
\end{figure*}

\subsubsection{Viewpoint-Augmented Planning}
\label{subsubsec:viewpoint_augmented_planning}
The sector-guided path provides global coverage over unknown regions. Finer exploration around each region is handled by frontier viewpoints in \(V^f\). After the unknown-region visiting order is determined, each frontier viewpoint is assigned to the nearest reachable unknown cluster center on the roadmap. The viewpoints are then inserted into the guided path without changing the region order. The insertion cost of viewpoint \(\mathbf{v}_i\) is
\begin{equation}
	\Delta c(\mathbf{v}_i,\mathbf{p}_a,\mathbf{p}_b)
	=
	\frac{
		c(\mathbf{p}_{a},\mathbf{v}_i)
		+
		c(\mathbf{v}_i,\mathbf{p}_{b})
		-
		c(\mathbf{p}_{a},\mathbf{p}_{b})
	}
	{g_i^{f}}.
	\label{eq:viewpoint_insertion_zh}
\end{equation}
where, \(\mathbf{p}_a\) and \(\mathbf{p}_b\) are the previous and next region points of \(\mathbf{v}_i\). If \(\mathbf{v}_i\) belongs to the region associated with path point \(\mathbf{p}_k\), only \(\Delta c(\mathbf{v}_i,\mathbf{p}_{k-1},\mathbf{p}_k)\) and \(\Delta c(\mathbf{v}_i,\mathbf{p}_k,\mathbf{p}_{k+1})\) are evaluated. The viewpoint is inserted into the lower-cost interval. If multiple viewpoints belong to the same interval, they are ordered by increasing insertion cost. As shown in Fig.~\ref{fig7}(c), the proposed method first preserves a region-level coverage direction and then supplements local observations inside each region. Compared with viewpoint-only TSP in Fig.~\ref{fig7}(a), it obtains stronger global consistency from unknown-region connectivity. Compared with the mixed TSP in Fig.~\ref{fig7}(b) , it reduces repeated revisits and produces a more coherent visiting order. The final exploration path is sent to the controller, and the system replans after a fixed time interval.

\begin{table}[t]
	\centering
	\caption{Parameters Setting}
	\label{tab:experiment_parameters}
	\footnotesize
	\renewcommand{\arraystretch}{1.08}
	\setlength{\tabcolsep}{2.4pt}
	\resizebox{\linewidth}{!}{%
		\begin{tabular}{clccc}
			\hline\hline
			\textbf{Type} & \textbf{Parameter} & \textbf{Section} & \textbf{Notation} & \textbf{Value} \\
			\hline
			\multirow{6}{*}{\shortstack{Ours\\Params}}
			& Cost map resolution & IV,V  & \(r_l\) & 0.25 \\
			& Ground threshold & IV-A & \(\eta_{\max},u_{\max}\) &  \(30^\circ, 0.8\) \\
			& Height threshold& IV-B & \(h_{\mathrm{trav}},h_{\mathrm{occ}}\) & 0.7m, 0.1m \\
			& Sliding window size & V-A & \(W^s\) & 15m \\
			& Clustering threshold  & V-C & \(d_c\) & 5m \\
			& Graph connect radius & VI-A & \(r_c\) & 3m \\
			\hline
			\multirow{4}{*}{\shortstack{Benchmark\\Common\\Params}}
			& Max linear velocity & -- & -- & \(2.0~\mathrm{m/s}\) \\
			& Max angular velocity & -- & -- & \(1.57~\mathrm{rad/s}\) \\
			& Sensor model & -- & -- &Velodyne VLP16 \\
			& Sensor range & -- & -- & \(10~\mathrm{m}\) \\
			\hline\hline
		\end{tabular}
	}
	\vspace{-0.4cm}
\end{table}
\section{Simulation Benchmark Results and Analysis}
\label{sec:simulation_benchmark}
To ensure consistent and reproducible benchmarking and ablation studies, we developed a lightweight point-cloud-based autonomous exploration simulator for ground robots. Seven scenes were selected to cover varying branch complexity, terrain roughness, and planning difficulty: Classical Parking, Large Maze, Small Terrain, Complex Terrain, Uneven Cave, Uneven Forest, and Rugged Hill. These scenes include narrow passages, complex connectivity, open outdoor areas, and large-scale terrains with significant elevation variation, enabling comprehensive evaluation of efficiency, completeness, and robustness. The environments and simulator are shown in Fig.~\ref{fig:robo}. Table~\ref{tab:experiment_parameters} summarizes the representative method-specific parameters used in Sections~\ref{sec:terrain_representation}--\ref{sec:roadmap_exploration_planning}, as well as the benchmark settings for all planners. Exploration is considered complete when the achieved coverage exceeds 95\%.

We compare the proposed method with six state-of-the-art exploration planners: GPB~\cite{22}, TARE~\cite{11}, LRAE~\cite{24}, FAEL~\cite{17}, STG~\cite{26}, and UFEP~\cite{25}. FAEL and STG are mainly designed for relatively flat structured environments, whereas TARE, LRAE, GPB, and UFEP provide varying capability on uneven terrain. Performance is evaluated using exploration time, traveled distance, final coverage, and average velocity. Each planner is tested 10 times in each scene, and the statistics are reported in Table~\ref{tab:baseline_exploration_statistics}. Fig.~\ref{fig:coverage_curves} shows the coverage--time curve of the run with the highest final coverage for each method. All simulations are conducted on Ubuntu 20.04 with an Intel Core i7-14700K CPU, an NVIDIA GeForce RTX 3090 24-GB GPU, and 128 GB RAM.

\begin{table*}[t]
	\centering
	\caption{Exploration Statistics. Best results in \colorbox{bestcyan}{cyan}, second-best in \colorbox{secondviolet}{violet}; failed runs in \colorbox{failedgray}{gray}\,.}
	\label{tab:baseline_exploration_statistics}
	\setlength{\tabcolsep}{2.5pt}
	\renewcommand{\arraystretch}{0.95}
	\resizebox{\textwidth}{!}{%
		\begin{tabular}{cM|FFFF|FFFF|FFFF|FFFN}
			\hline\hline
			\multirow{2}{*}{\textbf{Scene}} & \multirow{2}{*}{\textbf{Method}}
			& \multicolumn{4}{c|}{\textbf{Exploration Time (s)} $\downarrow$}
			& \multicolumn{4}{c|}{\textbf{Movement Distance (m)} $\downarrow$}
			& \multicolumn{4}{c|}{\textbf{Coverage (m$^2$)} $\uparrow$}
			& \multicolumn{4}{c}{\textbf{Avg Velocity (m/s)}$\uparrow$} \\
			&  & \textbf{Avg} & \textbf{Std} & \textbf{Max} & \textbf{Min} & \textbf{Avg} & \textbf{Std} & \textbf{Max} & \textbf{Min} & \textbf{Avg} & \textbf{Std} & \textbf{Max} & \textbf{Min} & \textbf{Avg} & \textbf{Std} & \textbf{Max} & \textbf{Min} \\
			\hline
			\multirow{7}{*}{\scenename{Classical}{Parking}} & GPB & 977.2 & 73.60 & 1000.0 & 767.7 & 801.1 & 61.30 & 844.9 & 646.9 & 3620.9 & 382.10 & 4032.9 & 2776.2 & 0.76 & 0.02 & 0.79 & 0.72 \\
			& TARE & 679.7 & 192.90 & 1000.0 & 457.7 & 788.3 & 276.20 & 1116.8 & 306.4 & 4008.8 & 665.30 & 4494.2 & 2699.5 & 1.28 & 0.39 & 1.89 & 0.77 \\
			& LRAE & 787.1 & 281.50 & 1000.0 & 385.9 & \bestavg{378.4} & 107.20 & 563.0 & 256.1 & 3007.2 & 906.10 & 4002.5 & 1335.4 & 0.38 & 0.12 & 0.57 & 0.26 \\
			& FAEL & \secondavg{328.9} & 26.60 & 383.4 & 299.0 & 442.3 & 44.80 & 540.2 & 392.1 & 4607.8 & 38.30 & 4621.6 & 4602.1 & 1.29 & 0.06 & 1.38 & 1.22 \\
			& STG & 344.9 & 19.90 & 389.1 & 318.0 & 474.2 & 28.80 & 547.1 & 449.5 & \secondavg{4613.6} & 10.40 & 4626.7 & 4602.4 & \secondavg{1.33} & 0.06 & 1.40 & 1.23 \\
			\failrow
			& UFEP & 1000.0 & 0.00 & 1000.0 & 1000.0 & 245.3 & 66.10 & 312.1 & 180.0 & 1937.0 & 602.30 & 2298.8 & 1241.7 & 0.31 & 0.07 & 0.37 & 0.22 \\
			& OURS & \bestavg{264.5} & 8.37 & 274.1 & 249.9 & \secondavg{392.5} & 18.10 & 413.7 & 362.4 & \bestavg{4615.3} & 9.23 & 4632.1 & 4603.1 & \bestavg{1.45} & 0.05 & 1.52 & 1.38 \\
			\hline
			\multirow{7}{*}{\scenename{Large}{Maze}} & GPB & 1710.6 & 340.60 & 2000.0 & 1335.6 & 1168.9 & 778.10 & 1630.2 & 270.6 & 6019.4 & 4051.80 & 8667.8 & 1355.0 & 0.67 & 0.24 & 0.85 & 0.40 \\
			& TARE & 901.5 & 88.10 & 1054.7 & 787.8 & 1425.6 & 88.90 & 1564.6 & 1263.1 & 9184.6 & 82.40 & 9293.9 & 9048.6 & \secondavg{1.57} & 0.09 & 1.69 & 1.41 \\
			& LRAE & 1643.4 & 482.20 & 2000.0 & 868.9 & 1294.8 & 305.70 & 1669.2 & 932.7 & 8427.7 & 838.30 & 9217.0 & 7247.5 & 1.32 & 0.14 & 1.42 & 1.12 \\
			& FAEL & \secondavg{751.5} & 47.60 & 801.4 & 667.1 & 1204.9 & 93.30 & 1294.3 & 1049.7 & \bestavg{9221.4} & 16.30 & 9255.8 & 9206.3 & \secondavg{1.57} & 0.03 & 1.62 & 1.50 \\
			& STG & 1257.1 & 643.40 & 2000.0 & 679.2 & \bestavg{966.3} & 299.30 & 1458.5 & 597.3 & 7369.6 & 1941.60 & 9256.2 & 4576.0 & 1.54 & 0.08 & 1.67 & 1.43 \\
			\failrow
			& UFEP & 2000.0 & 0.00 & 2000.0 & 2000.0 & 191.7 & 149.40 & 486.7 & 93.4 & 1463.9 & 897.00 & 3411.5 & 955.9 & 0.32 & 0.04 & 0.42 & 0.27 \\
			& OURS & \bestavg{631.4} & 22.70 & 678.4 & 603.3 & \secondavg{1112.8} & 38.00 & 1198.2 & 1061.6 & \secondavg{9218.7} & 54.70 & 9328.7 & 9103.4 & \bestavg{1.75} & 0.02 & 1.78 & 1.72 \\
			\hline
			\multirow{7}{*}{\scenename{Small}{Terrain}} & GPB & 313.1 & 99.90 & 471.9 & 161.0 & 177.9 & 75.50 & 310.8 & 76.2 & \secondavg{1059.5} & 44.30 & 1080.2 & 935.7 & 0.69 & 0.12 & 0.96 & 0.53 \\
			\failrow
			& TARE & 500.0 & 0.00 & 500.0 & 500.0 & 36.5 & 25.50 & 90.6 & 22.2 & 539.4 & 98.20 & 754.3 & 476.2 & 0.23 & 0.14 & 0.54 & 0.15 \\
			& LRAE & \secondavg{113.9} & 32.50 & 158.9 & 79.0 & \secondavg{131.5} & 47.00 & 187.8 & 43.9 & 1059.3 & 54.20 & 1081.1 & 936.3 & \secondavg{1.19} & 0.42 & 1.55 & 0.30 \\
			\failrow
			& FAEL & 500.0 & 0.00 & 500.0 & 500.0 & 87.5 & 73.10 & 223.6 & 10.4 & 603.7 & 181.00 & 822.6 & 324.4 & 0.27 & 0.16 & 0.49 & 0.05 \\
			\failrow
			& STG & 500.0 & 0.00 & 500.0 & 500.0 & 15.5 & 9.40 & 33.8 & 8.6 & 444.0 & 45.90 & 570.6 & 408.0 & 0.07 & 0.07 & 0.20 & 0.01 \\
			\failrow
			& UFEP & 500.0 & 0.00 & 500.0 & 500.0 & 102.4 & 82.10 & 316.2 & 22.4 & 688.3 & 106.70 & 915.5 & 558.3 & 0.51 & 0.12 & 0.68 & 0.32 \\
			& OURS & \bestavg{58.6} & 6.17 & 65.0 & 47.0 & \bestavg{88.9} & 11.80 & 98.5 & 67.6 & \bestavg{1080.1} & 0.82 & 1081.4 & 1079.2 & \bestavg{1.48} & 0.07 & 1.58 & 1.37 \\
			\hline
			\multirow{7}{*}{\scenename{Complex}{Terrain}} & GPB & 389.7 & 104.80 & 640.0 & 261.8 & 269.7 & 66.20 & 438.3 & 204.8 & \secondavg{3478.6} & 62.60 & 3511.9 & 3359.1 & 0.75 & 0.12 & 0.86 & 0.49 \\
			& TARE & 358.7 & 130.70 & 582.6 & 223.9 & \bestavg{222.6} & 26.30 & 271.4 & 201.8 & 3289.3 & 102.80 & 3498.3 & 3150.6 & 0.65 & 0.19 & 0.90 & 0.33 \\
			& LRAE & \secondavg{185.9} & 46.00 & 303.9 & 136.9 & 258.6 & 51.80 & 317.9 & 133.7 & 3371.1 & 441.80 & 3519.2 & 2113.7 & \secondavg{1.44} & 0.38 & 1.63 & 0.37 \\
			\failrow
			& FAEL & 863.1 & 290.30 & 1000.0 & 248.0 & 76.9 & 67.10 & 224.4 & 24.5 & 1218.9 & 876.30 & 3107.7 & 622.7 & 0.21 & 0.19 & 0.52 & 0.04 \\
			\failrow
			& STG & 714.2 & 351.30 & 1000.0 & 226.9 & 225.9 & 125.20 & 442.1 & 68.1 & 2282.6 & 911.30 & 3250.4 & 964.8 & 0.78 & 0.41 & 1.72 & 0.37 \\
			\failrow
			& UFEP & 1000.0 & 0.00 & 1000.0 & 1000.0 & 178.9 & 132.50 & 356.7 & 54.6 & 1392.7 & 626.20 & 2462.8 & 699.6 & 0.41 & 0.06 & 0.53 & 0.35 \\
			& OURS & \bestavg{141.7} & 10.30 & 154.2 & 125.9 & \secondavg{228.0} & 19.10 & 258.5 & 193.8 & \bestavg{3519.1} & 10.10 & 3537.3 & 3508.5 & \bestavg{1.61} & 0.08 & 1.73 & 1.47 \\
			\hline
			\multirow{7}{*}{\scenename{Uneven}{Cave}} & GPB & 2266.2 & 545.40 & 3000.0 & 1424.7 & 2615.5 & 949.20 & 4247.1 & 576.0 & 10932.2 & 2897.30 & 14721.0 & 4737.3 & 1.09 & 0.07 & 1.16 & 0.92 \\
			& TARE & \secondavg{1899.9} & 947.20 & 3000.0 & 1093.8 & \bestavg{1594.8} & 240.40 & 1831.6 & 1269.2 & \secondavg{11665.9} & 1604.80 & 12982.5 & 9559.3 & \secondavg{1.63} & 0.05 & 1.69 & 1.48 \\
			\failrow
			& LRAE & 3000.0 & 0.00 & 3000.0 & 3000.0 & 182.1 & 41.40 & 251.8 & 110.3 & 1522.0 & 25.80 & 1564.6 & 1473.1 & 0.59 & 0.13 & 0.83 & 0.39 \\
			\failrow
			& FAEL & 3000.0 & 0.00 & 3000.0 & 3000.0 & 2426.3 & 965.70 & 4310.3 & 721.4 & 4226.9 & 1539.70 & 7126.0 & 2990.9 & 0.88 & 0.22 & 1.44 & 0.69 \\
			\failrow
			& STG & 3000.0 & 0.00 & 3000.0 & 3000.0 & 103.4 & 61.70 & 152.5 & 10.1 & 2286.3 & 886.30 & 2860.7 & 975.9 & 0.33 & 0.18 & 0.47 & 0.05 \\
			\failrow
			& UFEP & 3000.0 & 0.00 & 3000.0 & 3000.0 & 412.0 & 246.30 & 767.3 & 129.1 & 2577.2 & 1428.10 & 4823.3 & 1270.1 & 0.41 & 0.05 & 0.48 & 0.33 \\
			& OURS & \bestavg{1000.9} & 57.00 & 1063.7 & 888.8 & \secondavg{1786.5} & 95.70 & 1888.3 & 1606.5 & \bestavg{15230.0} & 45.90 & 15348.0 & 15202.0 & \bestavg{1.77} & 0.02 & 1.80 & 1.74 \\
			\hline
			\multirow{7}{*}{\scenename{Uneven}{Forest}} & GPB & \secondavg{925.2} & 178.40 & 1000.0 & 427.1 & \secondavg{1067.9} & 266.80 & 1287.3 & 351.6 & \secondavg{5137.9} & 1001.30 & 5966.5 & 3097.8 & \secondavg{1.16} & 0.06 & 1.25 & 1.08 \\
			\failrow
			& TARE & 1000.0 & 0.00 & 1000.0 & 1000.0 & 14.3 & 14.00 & 48.2 & 1.4 & 994.5 & 300.70 & 1686.5 & 707.2 & 0.05 & 0.04 & 0.16 & 0.01 \\
			\failrow
			& LRAE & 1000.0 & 0.00 & 1000.0 & 1000.0 & 147.9 & 99.70 & 301.9 & 56.4 & 810.5 & 46.90 & 893.2 & 717.1 & 0.22 & 0.04 & 0.31 & 0.18 \\
			\failrow
			& FAEL & 1000.0 & 0.00 & 1000.0 & 1000.0 & 37.1 & 36.40 & 117.9 & 30.0 & 1027.9 & 284.20 & 1520.8 & 580.1 & 0.07 & 0.04 & 0.12 & 0.01 \\
			\failrow
			& STG & 1000.0 & 0.00 & 1000.0 & 1000.0 & 13.1 & 1.10 & 14.7 & 11.0 & 1043.5 & 21.10 & 1063.5 & 990.6 & 0.06 & 0.01 & 0.06 & 0.04 \\
			\failrow
			& UFEP & 1000.0 & 0.00 & 1000.0 & 1000.0 & 96.3 & 194.90 & 638.8 & 8.9 & 865.2 & 139.70 & 1100.1 & 693.7 & 0.42 & 0.07 & 0.52 & 0.25 \\
			& OURS & \bestavg{254.3} & 31.60 & 297.9 & 192.5 & \bestavg{356.1} & 49.10 & 400.3 & 259.5 & \bestavg{5923.2} & 41.20 & 6003.3 & 5888.4 & \bestavg{1.39} & 0.07 & 1.51 & 1.27 \\
			\hline
			\failrow
			\multirow{7}{*}{\scenename{Rugged}{Hill}} & GPB & 3522.6 & 581.10 & 4000.0 & 2529.1 & 3575.4 & 1391.20 & 5161.0 & 388.3 & 13777.6& 6507.30 & 20158.3 & 4476.9 & 0.91 & 0.12 & 1.01 & 0.58 \\
			\failrow
			& TARE & 4000.0 & 0.00 & 4000.0 & 4000.0 & 32.2 & 24.60 & 87.4 & 9.5 & 1473.4 & 604.50 & 2400.5 & 931.6 & 0.07 & 0.05 & 0.19 & 0.03 \\
			\failrow
			& LRAE & 3726.8 & 864.00 & 4000.0 & 2734.7 & 841.2 & 689.80 & 2078.4 & 130.5 & 7253.1 & 3568.80 & 14065.1 & 2788.2 & 0.37 & 0.12 & 0.52 & 0.15 \\
			\failrow
			& FAEL & 4000.0 & 0.00 & 4000.0 & 4000.0 & 49.8 & 38.80 & 106.3 & 5.9 & 1129.2 & 330.40 & 1690.3 & 731.6 & 0.10 & 0.06 & 0.18 & 0.02 \\
			\failrow
			& STG & 4000.0 & 0.00 & 4000.0 & 4000.0 & 12.0 & 2.80 & 13.8 & 3.9 & 1033.2 & 104.60 & 1078.3 & 736.8 & 0.04 & 0.01 & 0.04 & 0.01 \\
			\failrow
			& UFEP & 4000.0 & 0.00 & 4000.0 & 4000.0 & 21.1 & 16.90 & 54.1 & 2.8 & 1042.9 & 88.90 & 1142.1 & 888.9 & 0.52 & 0.29 & 1.11 & 0.29 \\
			& OURS & \bestavg{1702.6} & 141.80 & 1865.6 & 1486.5 & \bestavg{2410.7} & 219.80 & 2716.2 & 2119.1 & \bestavg{40584.4} & 17.50 & 40612.0 & 40562.2 & \bestavg{1.41} & 0.04 & 1.48 & 1.36 \\
			\hline\hline
		\end{tabular}%
	}
	\vspace{-0.4cm}
\end{table*}

\subsection{Planner Performance Evaluation}
\label{subsec:planner_benchmark}
\subsubsection{Classical Maze-Like Scenes}
\label{subsubsec:classic}
\begin{figure*}[t]
	\centering
	\includegraphics[width=0.96\textwidth]{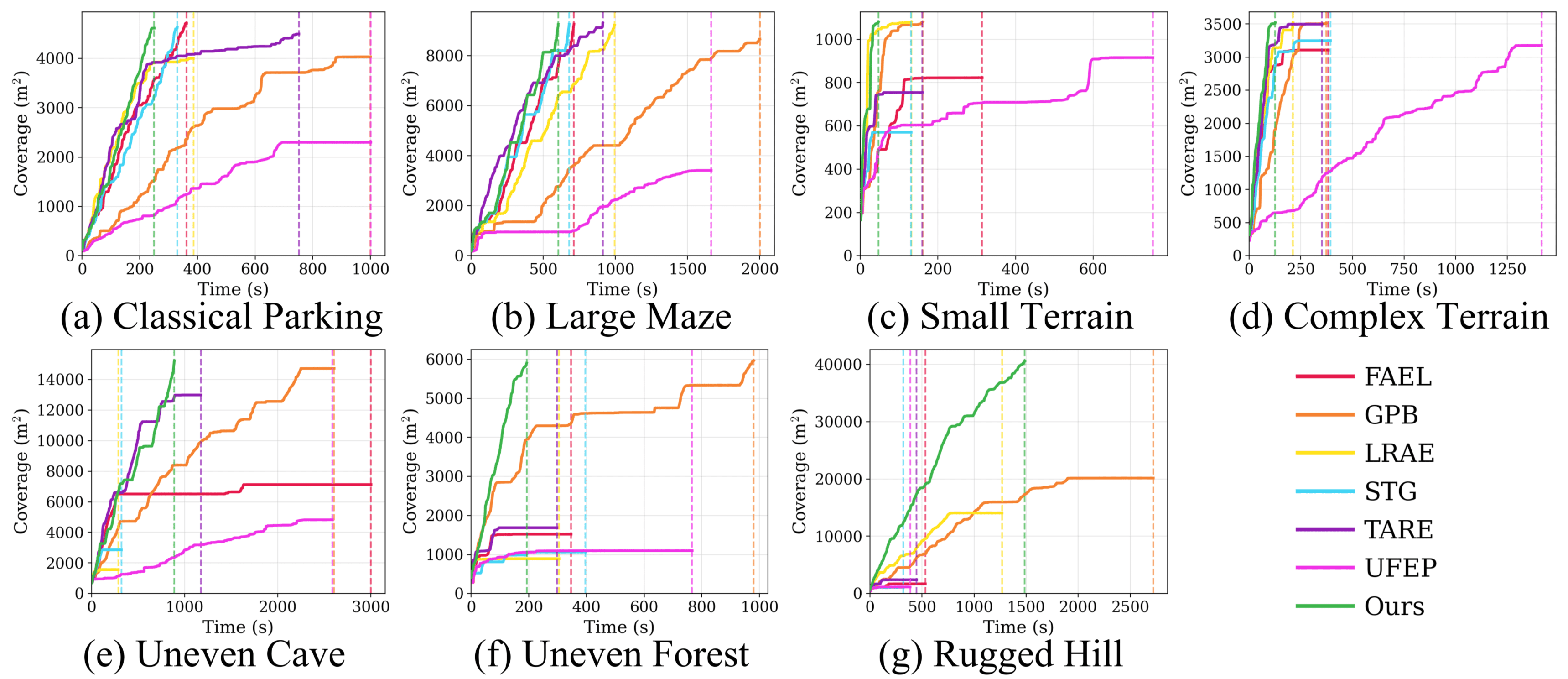}
	\caption{Coverage-over-time curves of different methods in different scenes.}
	\label{fig:coverage_curves}
	\vspace{-0.4cm}
\end{figure*}

Classical Parking and Large Maze contain narrow corridors, enclosed rooms, and multi-level branches. They evaluate global planning ability. The results are shown in Table~\ref{tab:baseline_exploration_statistics} and Fig.~\ref{fig:coverage_curves}(a),(b). TARE and LRAE tend to visit large unknown regions first. They can leave the start area quickly and move toward distant space in the early stage. However, when many short branches and small rooms exist, this global-greedy behavior skips local branches. The robot then spends more time revisiting them later, and coverage can remain incomplete. FAEL and STG rely more on local frontiers or skeleton structures. They handle nearby unknown boundaries stably and achieve high coverage in these planar structured scenes. However, without a unified visiting order over distant unknown regions, the robot often switches repeatedly between adjacent corridors, especially in Large Maze. GPB and UFEP use more conservative path search and traversability reasoning, but their planning frameworks are redundant and slow down exploration.

In contrast, the proposed method uses clustered sector unknown regions as global visiting targets and inserts frontier viewpoints after the regional order is determined. Region-level targets maintain a consistent forward direction and avoid short-sighted switching among adjacent frontiers. Viewpoint insertion preserves observation opportunities in narrow branches and small boundaries. Quantitatively, the proposed method achieves average exploration times of 264.5 s and 631.4 s in the two scenes. While keeping coverage near the best level, it reduces exploration time by 19.6\% and 16.0\% compared with the second-best method.

\begin{figure*}[t]
	\centering
	\includegraphics[width=0.96\textwidth]{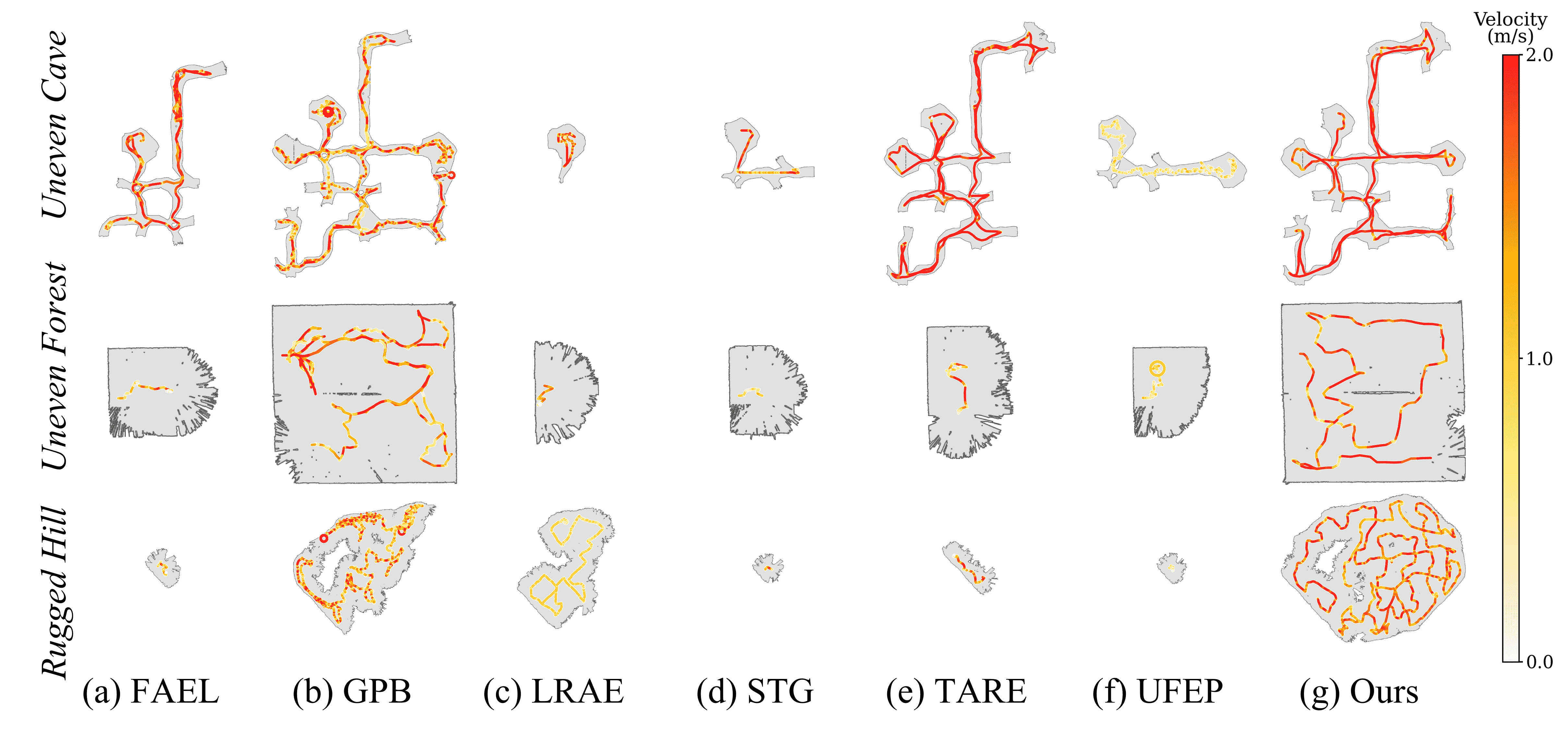}
	\caption{Trajectories of different methods in three challenging scenes.}
	\label{fig:trajectory_comparison}
	\vspace{-0.4cm}
\end{figure*}

\subsubsection{Uneven Open Environments}
\label{subsubsec:uneven}
Small Terrain and Complex Terrain are open uneven outdoor environments. They test exploration under terrain variation and local obstacles. Compared with maze-like scenes, these environments have fewer topological branches, but two-dimensional distance no longer reflects true motion cost. A planner must evaluate both whether a target is worth exploring and whether the path to it is safe. Table~\ref{tab:baseline_exploration_statistics} and Fig.~\ref{fig:coverage_curves}(c),(d) show that FAEL and STG, which rely on two-dimensional motion assumptions, often fail to detect frontiers or update skeletons on slopes. Most runs therefore do not complete exploration. TARE uses a simple height-difference constraint and can move in mild-slope regions, but it stalls near the steep hill in Small Terrain. UFEP uses a multi-layer elevation map to analyze terrain, but its exploration framework remains redundant and cannot maintain high-frequency planning. GPB directly analyzes traversability with three-dimensional voxels, but its planning is conservative and requires long time to cover open terrain. LRAE is designed for uneven open areas and usually moves quickly. However, it mainly uses large-region information to greedily choose the next target. Small unexplored areas can be ignored, causing repeated backtracking late in exploration.

The proposed method assigns terrain costs to roadmap paths through hierarchical traversability analysis. Its region gain considers both angular breadth and radial depth, allowing the robot to move quickly along low-cost terrain while still considering distant promising unknown space. In Small Terrain, the proposed method completes exploration in 58.6 s, reducing time by 48.5\% compared with the second-best method. In Complex Terrain, the average exploration time is 141.7 s, reducing time by 23.8\%.
\begin{figure*}[t]
	\centering
	\includegraphics[width=0.96\textwidth]{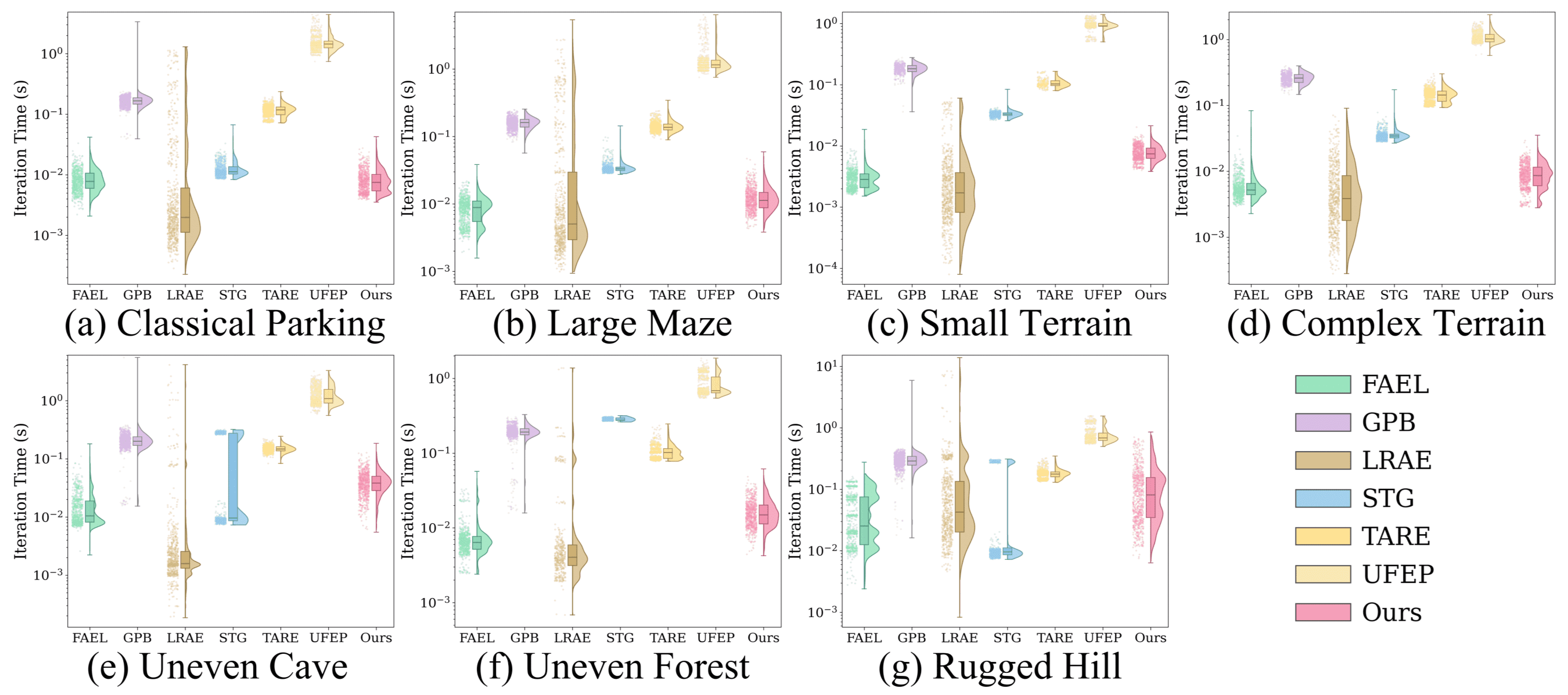}
	\caption{Algorithm runtime of different methods in different scenes.}
	\label{fig:planner_runtime}
	\vspace{-0.4cm}
\end{figure*}

\subsubsection{Large-Scale Complex Environments}
\label{subsubsec:complex}
Uneven Cave, Uneven Forest, and Rugged Hill jointly increase environment scale, terrain complexity, and topological uncertainty. They provide a comprehensive test of planner capability. Fig.~\ref{fig:trajectory_comparison} shows the trajectories of different methods in these three environments.

Uneven Cave contains long cave passages, dense branches, gravel surfaces, and ramps. It was originally used in aerial exploration evaluation. When used for ground robots, the planner must maintain large-scale topological connectivity while continuously rejecting targets that are unsuitable for ground motion. Except for the proposed method, only GPB and TARE explore large parts of this scene. TARE reaches an average coverage of 11665.9 m$^2$, but it is limited by terrain reachability near complex slopes and gravel regions. Its exploration time also has high variance. GPB maintains safety through three-dimensional voxel analysis, but its average exploration time reaches 2266.2 s and its trajectories contain inefficient detours. The proposed method combines reliable traversability analysis with efficient planning. It completes exploration in 1000.9 s and reaches 15230.0 m$^2$ coverage, reducing exploration time by 47.3\% and increasing coverage by 30.5\% compared with the second-best baseline.

Uneven Forest is constructed by stitching several real collected scenes. It contains dense trunks, narrow gaps, and undulating ground. Its main difficulty is that traversable gaps between trees are narrow. Local obstacles and ground undulation are easily confused in a two-dimensional projection, causing most methods to fail to generate executable targets. TARE, LRAE, FAEL, STG, and UFEP all stall in small regions. GPB is the only baseline that continues to explore because of its fine three-dimensional voxel analysis. However, without region-level global guidance, it frequently backtracks along explored passages in the forest. It requires 925.2 s on average and travels 1067.9 m. The proposed 8-bit obstacle encoding distinguishes trunk obstacles from continuous ground more stably. Sector segmentation organizes unknown space behind dense occlusion into compact targets, allowing the planner to maintain a clear direction through narrow passages. The proposed method completes exploration in 254.3 s, reducing time by 72.5\% and movement distance by 66.7\% relative to GPB, while increasing coverage to 5923.2 m$^2$.

Rugged Hill is the largest comprehensive scene. It is built from UAV mapping data and includes mountains, roads, town edges, and forested areas. It tests generalization, long-term operation, and roadmap scale control. Most baselines cannot start or sustain effective exploration. LRAE and GPB explore for some time, but planning cost grows as the explored area expands. They also lack stable global connectivity when crossing different terrain types. Their average coverages are only 7253.1 m$^2$ and 13777.6 m$^2$. The proposed method maintains dynamic roadmap updates and sparsification, so remote unknown regions contribute to path estimation without causing uncontrolled planning cost. Sector guidance also reduces ineffective detours when crossing hills and forests. The proposed method completes exploration in 1702.6 s on average, reducing time by 51.7\% compared with GPB. Its coverage reaches 40584.4 m$^2$, about 2.95 times that of GPB.

\subsubsection{Planning Efficiency}
\label{subsubsec:efficient}
Fig.~\ref{fig:planner_runtime} reports per-iteration planning runtime over 10 repeated experiments in seven scenes. The violin plots show runtime distributions, the scatter points show individual replanning records, and the box plots show medians and quartiles. The proposed method has a concentrated runtime distribution with small variance. In most scenes, the per-iteration runtime stays below 100 ms. Even in Rugged Hill, no obvious long tail appears.

This efficiency comes from three scale-control mechanisms. First, sector segmentation and history-aware clustering compress many grid-level unknown boundaries into a small number of region-level targets. The planner optimizes the order of unknown cluster centers rather than all frontiers or viewpoints. Second, the dynamic roadmap uses a near-dense, far-sparse maintenance strategy. Dense vertices near the robot support safe path queries in narrow spaces. Remote areas are compressed into a sparse topological skeleton through variable-voxel merging, redundant-edge migration, and bridge restoration. As the explored area grows, A* queries remain on a compact graph rather than degenerating into dense full-map search. Third, unknown regions and frontier viewpoints are not mixed into a single TSP. The planner first generates a sector-guided path and then inserts relevant frontier viewpoints without breaking the regional order. This hierarchical organization decouples global coverage decisions from local supplementary observations. It reduces combinatorial optimization cost and avoids repeated path evaluations among neighboring viewpoints.

\begin{figure*}[t]
	\centering
	\includegraphics[width=0.96\textwidth]{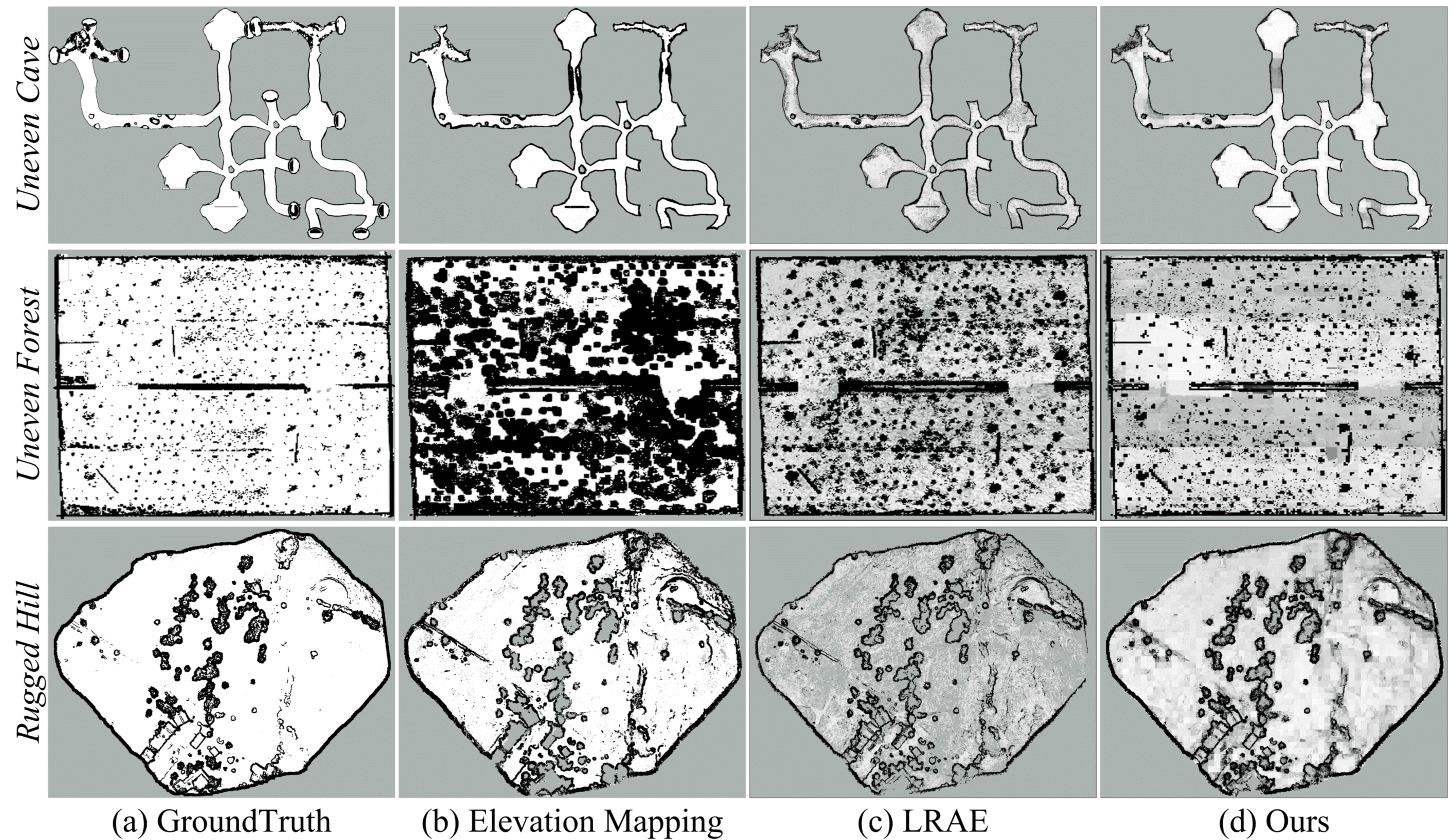}
	\caption{Comparison of ground traversability analysis results.}
	\label{fig:ground_segmentation}
	\vspace{-0.4cm}
\end{figure*}

\begin{table}[t]
	\centering
	\caption{Ground Segmentation Performance Comparison. Best results in \colorbox{bestcyan}{cyan}, second-best in \colorbox{secondviolet}{violet}\,.}
	\label{tab:ground_performance}
	\setlength{\tabcolsep}{2.5pt}
	\renewcommand{\arraystretch}{1.15}
	\resizebox{0.98\linewidth}{!}{%
		\begin{tabular}{llccccc}
			\hline\hline
			\textbf{Scene} & \textbf{Method} & \textbf{mIoU} $\uparrow$ & \textbf{Trav.\ IoU} $\uparrow$ & \textbf{Cov.\ Ratio} $\uparrow$ &\textbf{ Acc.\ Ratio} $\uparrow$ &\textbf{ RT (ms)} $\downarrow$ \\
			\midrule
			\multirow{3}{*}{\scenename{Uneven}{Cave}}
			& GPU-EM    & 0.532 & \secondavg{0.751} & \secondavg{0.839} & \secondavg{0.710} & 184.93 \\
			& LRAE   & \secondavg{0.544} & 0.714 & 0.754 & 0.684 & \secondavg{73.76} \\
			& Ours   & \bestavg{0.603} & \bestavg{0.842} & \bestavg{0.883} & \bestavg{0.804} & \bestavg{11.30} \\
			\midrule
			\multirow{3}{*}{\scenename{Uneven}{Forest}}
			& GPU-EM    & 0.303 & 0.432 & \secondavg{0.977} & 0.487 & 117.78 \\
			& LRAE   & \secondavg{0.567} & \secondavg{0.776} & 0.908 & \secondavg{0.781} & \secondavg{77.79} \\
			& Ours   & \bestavg{0.581} & \bestavg{0.886} & \bestavg{0.997} & \bestavg{0.889} & \bestavg{17.86} \\
			\midrule
			\multirow{3}{*}{\scenename{Rugged}{Hill}}
			& GPU-EM    & \secondavg{0.615} & \secondavg{0.815} & \secondavg{0.886} & \secondavg{0.790} & \secondavg{136.61} \\
			& LRAE   & 0.504 & 0.538 & 0.592 & 0.548 & 167.82 \\
			& Ours   & \bestavg{0.622} & \bestavg{0.875} & \bestavg{0.981} & \bestavg{0.874} & \bestavg{16.70} \\
			\hline\hline
		\end{tabular}
	}
	\vspace{-0.4cm}
\end{table}

\subsection{Ground Traversability Analysis Evaluation}
\label{subsec:ground_benchmark}
To validate the suitability of the hierarchical traversability analysis for exploration planning, we evaluate traversability analysis in Uneven Cave, Uneven Forest, and Rugged Hill. The proposed method is compared with GPU-based elevation mapping \cite{44} and LRAE \cite{24}. All methods incrementally build terrain maps along the same trajectory, avoiding path differences in the segmentation evaluation. The mIoU measures overall region classification quality. Trav.\ IoU focuses on traversable-region recognition. Cov.\ Ratio and Acc.\ Ratio are designed for autonomous exploration. Cov.\ Ratio measures how much of the true traversable region is recovered by the constructed map, reflecting whether many holes remain in traversable space. Acc.\ Ratio measures the accuracy of traversable regions, indicating whether obstacles and non-traversable areas are reliably excluded.  RT denotes per-frame processing time.

Fig.~\ref{fig:ground_segmentation} compares the constructed traversability cost maps. GPU-EM analyzes elevation maps and performs well in relatively open terrain such as Rugged Hill. In Uneven Cave and Uneven Forest, however, a single height representation tends to project cave ceilings, tree canopies, and other suspended structures onto the ground layer. This mixes obstacle boundaries with true traversable ground. LRAE uses fine-grained voxels to improve local geometry representation, but sparse point clouds and per-voxel decisions leave many unknown holes and noisy cells inside traversable regions. These false unknown areas can mislead target selection during exploration.

In contrast, the proposed method uses variable voxels for ground fitting. Open smooth areas are restored quickly by larger planes, reducing holes inside traversable space. Obstacle boundaries, steps, and terrain discontinuities automatically retain finer plane resolution. The 8-bit height code then records vertical occupancy inside each cell at low cost, allowing the system to distinguish continuous slopes, tree trunks, rocks, and suspended occlusions. The resulting terrain map has fewer holes, clearer obstacle boundaries, and low computation cost. Table~\ref{tab:ground_performance} shows that the proposed method achieves the best traversability analysis in all three scenes while maintaining the lowest runtime. On average, Trav.\ IoU, Cov.\ Ratio, and Acc.\ Ratio are 0.868, 0.954, and 0.856. They improve over the best baseline averages by 0.087, 0.053, and 0.096, respectively. The average per-frame runtime is 15.29 ms, while the fastest baseline average is 96.05 ms, giving an approximately 6.3 times speedup.

\begin{table*}[t]
	\centering
	\caption{Ablation Study -- Exploration Statistics. Best results in \colorbox{bestcyan}{cyan}, second-best in \colorbox{secondviolet}{violet}\,.}
	\label{tab:ablation_exploration_statistics}
	\footnotesize
	\setlength{\tabcolsep}{2.5pt}
	\renewcommand{\arraystretch}{0.95}
	\resizebox{0.92\textwidth}{!}{%
		\begin{tabular}{cl|cccc|cccc|cccc|cccc}
			\hline\hline
			\multirow{2}{*}{\textbf{Scene}} & \multirow{2}{*}{\textbf{Method}}
			& \multicolumn{4}{c|}{\textbf{Exploration Time (s)}$\downarrow$}
			& \multicolumn{4}{c|}{\textbf{Movement Distance (m)}$\downarrow$}
			& \multicolumn{4}{c|}{\textbf{Avg Velocity (m/s)}$\uparrow$}
			& \multicolumn{4}{c}{\textbf{Run Time (ms)}$\downarrow$} \\
			&  & \textbf{Avg} & \textbf{Std} & \textbf{Max} & \textbf{Min} & \textbf{Avg} & \textbf{Std} & \textbf{Max} & \textbf{Min} & \textbf{Avg} & \textbf{Std} & \textbf{Max} & \textbf{Min} & \textbf{Avg} & \textbf{Std} & \textbf{Max} & \textbf{Min} \\
			\hline
			\multirow{7}{*}{\scenename{Classical}{Parking}}
			& w/o UF & 314.5 & 19.40 & 355.0 & 286.9 & 435.6 & 25.50 & 466.3 & 402.5 & 1.34 & 0.04 & 1.39 & 1.29 & 13.5 & 0.97 & 14.9 & 12.4 \\
			& EuC & 299.6 & 33.30 & 354.0 & 263.9 & 430.5 & 47.10 & 506.7 & 381.6 & 1.40 & 0.03 & 1.43 & 1.35 & 8.6 & 0.70 & 9.5 & 7.8 \\
			& w/o UV & 284.3 & 30.60 & 329.5 & 256.0 & 428.8 & 43.50 & 499.8 & 397.2 & \bestavg{1.47} & 0.06 & 1.54 & 1.39 & \bestavg{7.1} & 0.10 & 7.3 & 7.0 \\
			& w/o DF & \secondavg{280.4} & 27.30 & 322.9 & 255.0 & \secondavg{411.8} & 22.70 & 449.0 & 393.5 & \secondavg{1.45} & 0.15 & 1.53 & 1.18 & 13.5 & 2.60 & 16.4 & 9.5 \\
			& F-TSP & 396.3 & 106.60 & 582.0 & 323.9 & 499.1 & 52.60 & 570.0 & 444.4 & 1.28 & 0.31 & 1.47 & 0.74 & 8.3 & 5.90 & 18.9 & 5.3 \\
			& RF-TSP & 323.4 & 57.40 & 383.5 & 252.9 & 465.1 & 72.20 & 559.2 & 381.7 & 1.41 & 0.06 & 1.46 & 1.33 & \secondavg{8.0} & 0.60 & 9.1 & 7.3 \\
			& OURS & \bestavg{264.5} & 8.37 & 274.1 & 249.9 & \bestavg{392.5} & 18.10 & 413.7 & 362.4 & \secondavg{1.45} & 0.05 & 1.52 & 1.38 & 8.4 & 1.62 & 12.3 & 7.0 \\
			\hline
			\multirow{7}{*}{\scenename{Complex}{Terrain}}
			& w/o UF & 172.2 & 35.90 & 213.2 & 121.6 & 261.4 & 59.20 & 333.4 & 190.6 & 1.50 & 0.10 & 1.57 & 1.32 & 14.3 & 2.90 & 17.3 & 10.9 \\
			& EuC & \secondavg{142.4} & 13.90 & 159.1 & 126.1 & \bestavg{219.1} & 17.60 & 241.2 & 198.1 & 1.53 & 0.06 & 1.58 & 1.43 & 13.2 & 0.50 & 14.0 & 12.7 \\
			& w/o UV & 166.4 & 8.40 & 176.3 & 155.0 & 241.2 & 11.20 & 256.1 & 227.6 & 1.43 & 0.06 & 1.51 & 1.35 & \bestavg{9.5} & 1.00 & 10.6 & 8.0 \\
			& w/o DF & 148.9 & 15.60 & 168.7 & 134.0 & 231.6 & 18.60 & 251.8 & 203.0 & 1.55 & 0.10 & 1.71 & 1.45 & 19.6 & 1.60 & 22.1 & 18.2 \\
			& F-TSP & 156.8 & 8.30 & 167.1 & 148.1 & 266.3 & 21.20 & 295.6 & 243.8 & \bestavg{1.69} & 0.05 & 1.76 & 1.64 & 10.5 & 0.30 & 10.9 & 9.9 \\
			& RF-TSP & 149.7 & 12.50 & 161.9 & 130.0 & 232.3 & 15.40 & 241.3 & 205.0 & 1.54 & 0.05 & 1.59 & 1.46 & 30.3 & 1.30 & 31.8 & 28.2 \\
			& OURS & \bestavg{141.7} & 10.30 & 154.2 & 125.9 & \secondavg{228.0} & 19.10 & 258.5 & 193.8 & \secondavg{1.61} & 0.08 & 1.73 & 1.47 & \secondavg{9.6} & 1.40 & 11.4 & 6.5 \\
			\hline\hline
		\end{tabular}%
	}
	\vspace{-0.4cm}
\end{table*}

\begin{table}[t]
	\centering
	\caption{Ablation Study -- Ground Segmentation Performance. Best results in \colorbox{bestcyan}{cyan}, second-best in \colorbox{secondviolet}{violet}\,.}
	\label{tab:ablation_ground_performance}
	\setlength{\tabcolsep}{2.5pt}
	\renewcommand{\arraystretch}{1.15}
	\resizebox{0.98\linewidth}{!}{%
		\begin{tabular}{llccccc}
			\hline\hline
			\textbf{Scene} & \textbf{Method} & \textbf{mIoU} $\uparrow$ & \textbf{Trav.\ IoU} $\uparrow$ & \textbf{Cov.\ Ratio} $\uparrow$ & \textbf{Acc.\ Ratio} $\uparrow$ & \textbf{RT (ms)} $\downarrow$ \\
			\midrule
			\multirow{3}{*}{\scenename{Uneven}{Cave}}
			& w/o code  & \secondavg{0.496} & \secondavg{0.715} & \bestavg{0.896} & \secondavg{0.705} & \bestavg{10.80} \\
			& w/o voxel & 0.340 & 0.436 & 0.809 & 0.480 & 11.43 \\
			& Ours      & \bestavg{0.603} & \bestavg{0.842} & \secondavg{0.883} & \bestavg{0.804} & \secondavg{11.30} \\
			\midrule
			\multirow{3}{*}{\scenename{Uneven}{Forest}}
			& w/o code  & \secondavg{0.449} & \secondavg{0.704} & \bestavg{0.997} & \secondavg{0.722} & \secondavg{11.63} \\
			& w/o voxel & 0.099 & 0.041 & 0.779 & 0.154 & \bestavg{6.33} \\
			& Ours      & \bestavg{0.581} & \bestavg{0.886} & \bestavg{0.997} & \bestavg{0.889} & 17.86 \\
			\midrule
			\multirow{3}{*}{\scenename{Rugged}{Hill}}
			& w/o code  & 	\bestavg{0.643} & \bestavg{0.879} & \bestavg{0.982} & \bestavg{0.880} & \secondavg{12.52} \\
			& w/o voxel & 0.564 & 0.602 & 0.714 & 0.633 & \bestavg{11.55} \\
			& Ours      &  \secondavg{0.622} & \secondavg{0.875} & \secondavg{0.981} & \secondavg{0.874} & 16.70 \\
			\hline\hline
		\end{tabular}
	}
	\vspace{-0.4cm}
\end{table}

\section{Ablation Study}
\label{sec:ablation_study}
We conduct ablation studies on the exploration planning framework and the ground traversability analysis module. The ablations use the same simulator and metrics as the benchmark. Only the tested design component is changed, allowing its contribution to be observed directly.

\subsection{Planning Framework Ablation}
\label{subsec:planner_ablation}
For the planning framework, six variants are evaluated around unknown-region processing, roadmap updates, and path organization. The abbreviations in Table~\ref{tab:ablation_exploration_statistics} are:
\begin{itemize}
	\item \textbf{w/o UF}: removes the unknown-region filtering in Section~\ref{subsubsec:unknown_region_filtering}. All detected unknown regions are kept for planning.
	\item \textbf{EuC}: replaces unknown-region clustering in Section~\ref{subsubsec:unknown_region_clustering} with conventional Euclidean clustering, without historical inheritance or adjacency expansion.
	\item \textbf{w/o UV}: removes unknown-vertex maintenance in Section~\ref{subsubsec:unknown_vertex_updating}. The roadmap uses only known traversable vertices.
	\item \textbf{w/o DF}: removes roadmap dynamic management in Sections~\ref{subsubsec:local_graph_pruning} and~\ref{subsubsec:global_graph_merging}. Local pruning and global merging are disabled.
	\item \textbf{F-TSP}: uses only the frontier viewpoints extracted in Section~\ref{subsubsec:viewpoint_extraction} as TSP targets, without region-level guidance.
	\item \textbf{RF-TSP}: places region centers from Section~\ref{subsubsec:guided_path_generation} and frontier viewpoints from Section~\ref{subsubsec:viewpoint_augmented_planning} into the same TSP, replacing the proposed hierarchical path organization.
\end{itemize}

\subsubsection{Unknown-Region Processing Ablation}
Unknown-region processing mainly affects the quality of the target set. The w/o UF variant keeps all unknown sector regions. Its exploration time and per-iteration runtime both increase in the two scenes, because many low-quality fragmented pseudo-unknown regions are retained and repeatedly evaluated. Under EuC, the exploration time and movement distance in Classical Parking increase by 13.3\% and 9.7\%, respectively. The effect is smaller in Complex Terrain. This indicates that Euclidean clustering can reduce the number of unknown targets, but it relies only on spatial distance and lacks historical inheritance. In narrow passages, target centers can become unstable across consecutive frames. The proposed clustering is more suitable for online exploration because it compresses target scale while preserving unknown-region connectivity and temporal consistency.

\begin{figure*}[t]
	\centering
	\includegraphics[width=0.96\textwidth]{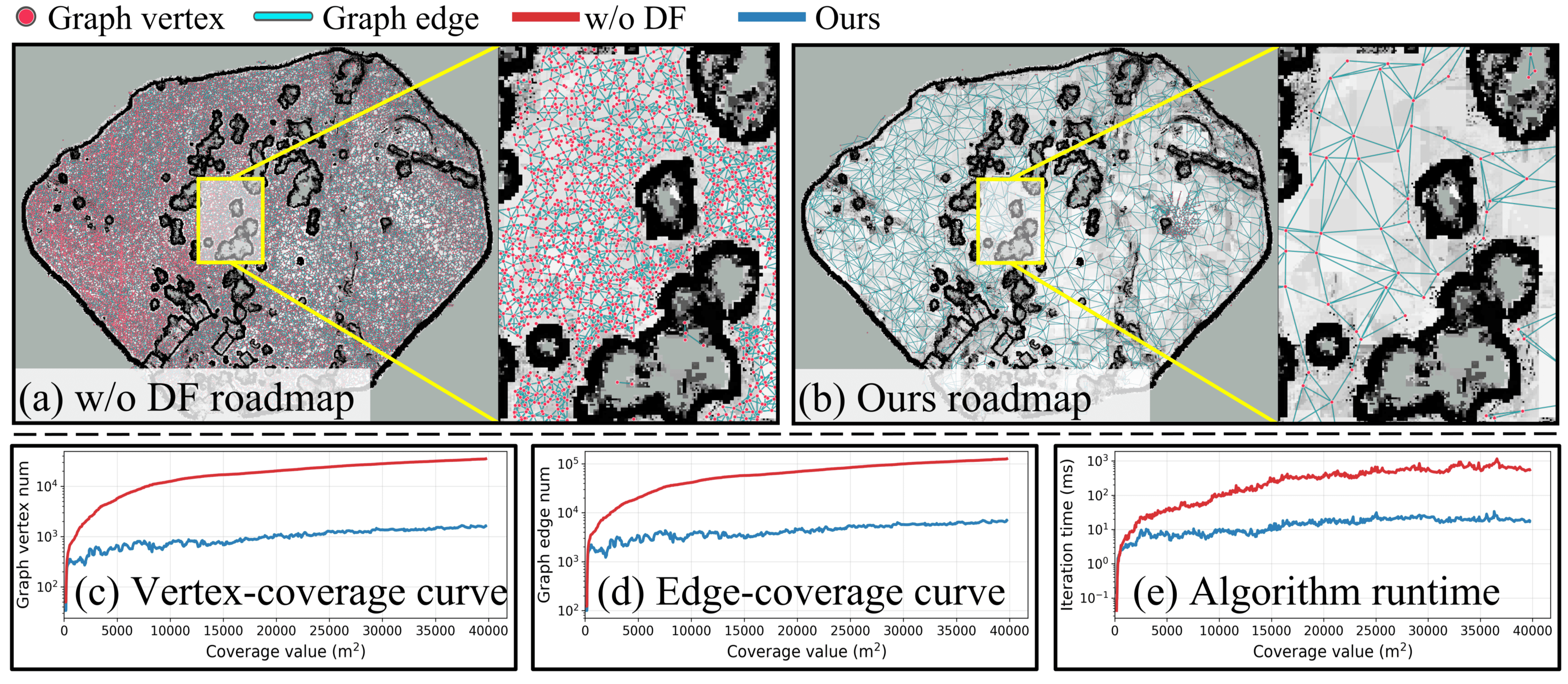}
	\caption{Roadmap sparsification. (a) Roadmap produced by w/o DF after exploration in Rugged Hill. (b) Roadmap produced by the proposed method. (c) Number of vertices versus covered area. (d) Number of edges versus covered area. (e) Graph update time versus covered area.}
	\label{fig:sparse}
	\vspace{-0.4cm}
\end{figure*}

\subsubsection{Roadmap Update Ablation}
Roadmap update ablations show the influence of topological representation on exploration planning. Without unknown vertices (w/o UV), the exploration time increases by 7.5\% and 17.4\% in the two scenes. Although removing unknown vertices slightly reduces graph-query cost, the planner can only expand through verified traversable vertices. It tends to stay near local boundaries and cannot estimate the potential connectivity between remote unknown regions and current known space. When roadmap dynamic management is disabled (w/o DF), runtime increases clearly. Fig.~\ref{fig:sparse}(a) and (b) compare the roadmaps of w/o DF and the proposed method after exploration in Rugged Hill. The proposed roadmap is visibly sparser. Fig.~\ref{fig:sparse}(c)--(e) show the number of vertices, number of edges, and update time as coverage grows. The sparsification strategy suppresses long-term accumulation of redundant nodes while maintaining stable update speed.

\subsubsection{Path-Organization Ablation}
Path-organization ablations reveal the hierarchy between region-level guidance and local viewpoint supplementation. F-TSP uses only frontier viewpoints as TSP targets. In Classical Parking, exploration time increases by 49.8\%, and movement distance increases by 27.2\%. In Complex Terrain, movement distance also increases by 16.8\%. Local viewpoints are useful for boundary observations, but they cannot control the global exploration direction alone. RF-TSP directly mixes region centers and frontier viewpoints into one TSP. Exploration time in Classical Parking increases by 22.3\%, because mixed targets cause frequent switching between regional expansion and local filling. Runtime in Complex Terrain rises to 30.3 ms, because open environments contain more targets with different spatial scales, enlarging the optimization problem. The proposed method first generates a sector-guided path and then inserts associated viewpoints, preserving both global exploration efficiency and local coverage completeness.

\begin{figure*}[t]
	\centering
	\includegraphics[width=0.96\textwidth]{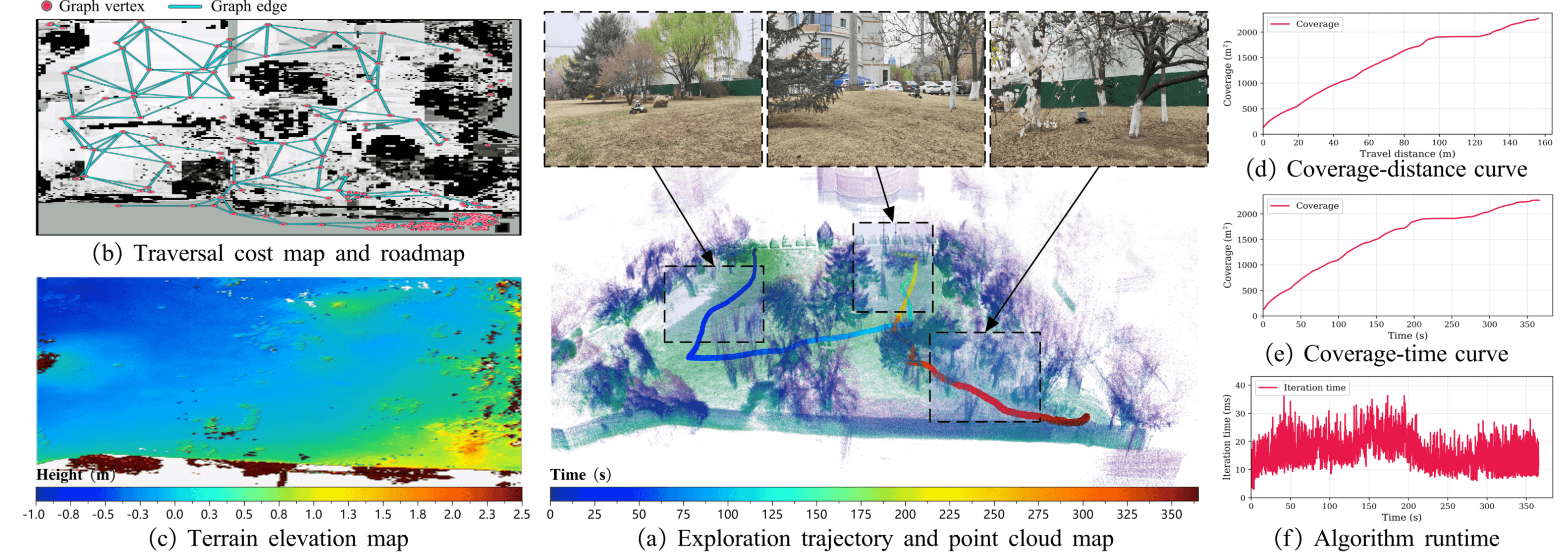}
	\caption{Small-scale field exploration in an open lawn scene. (a) Exploration trajectory. (b) Traversability cost map and roadmap after exploration. (c) Elevation map of the terrain. (d) Coverage versus movement distance. (e) Coverage versus runtime. (f) Algorithm iteration time versus runtime.}
	\label{fig:real_small}
	\vspace{-0.4cm}
\end{figure*}
\subsection{Ground Traversability Ablation}
\label{subsec:ground_ablation}

Table~\ref{tab:ablation_ground_performance} analyzes the two terrain-representation components in Section~\ref{sec:terrain_representation}. The w/o code variant removes the 8-bit height encoding in Section~\ref{subsec:height_obstacle_encoding} and judges obstacles only from height differences inside each grid cell. The w/o voxel variant removes variable-voxel ground fitting in Section~\ref{subsec:variable_voxel_ground_fitting} and fits planes with fixed-size voxels.

When 8-bit height encoding is removed, Cov.\ Ratio remains high. This indicates that variable voxels can recover most continuous ground. However, Trav.\ IoU and Acc.\ Ratio drop, especially in Uneven Forest, where they are 0.182 and 0.167 lower than the full method. Height difference alone cannot reliably distinguish continuous slopes, tree trunks, rocks, and suspended occlusions. It can incorrectly merge local obstacle structures into traversable ground. The complete method records obstacle occupancy distribution inside each cell and improves obstacle classification without a large computation increase.

With fixed-voxel ground fitting, performance drops more severely. The average Trav.\ IoU decreases by 0.509, and the Trav.\ IoU in Uneven Forest is only 0.041. A fixed voxel scale cannot adapt to both open continuous ground and dense obstacles. Large voxels smooth out local obstacles, while small voxels leave holes in sparse point clouds. The full method is slightly slower than the simplified variants, but RT remains between 11.30 ms and 17.86 ms, and segmentation performance is best or near best. Variable voxels therefore mainly guarantee complete traversable-region recovery, while 8-bit encoding mainly improves obstacle recognition.

\section{Real-World Experiments}
\label{sec:real_world_experiments}
To further validate real-world usability and cross-scene generalization, we conducted robot exploration experiments in three outdoor scenes with different scales and terrain complexity. The robot platform is a SCOUT MINI equipped with a Velodyne VLP-16 LiDAR for environment observation. IG-LIO\cite{45} provides real-time LiDAR-inertial odometry. The proposed exploration algorithm runs on an onboard NUC computer with an Intel Core i5-1240P CPU and 8 GB RAM. For safety, the maximum linear and angular velocities are limited to \(0.5~\mathrm{m/s}\) and \(1.0~\mathrm{rad/s}\), respectively.

\begin{figure*}[t]
	\centering
	\includegraphics[width=0.96\textwidth]{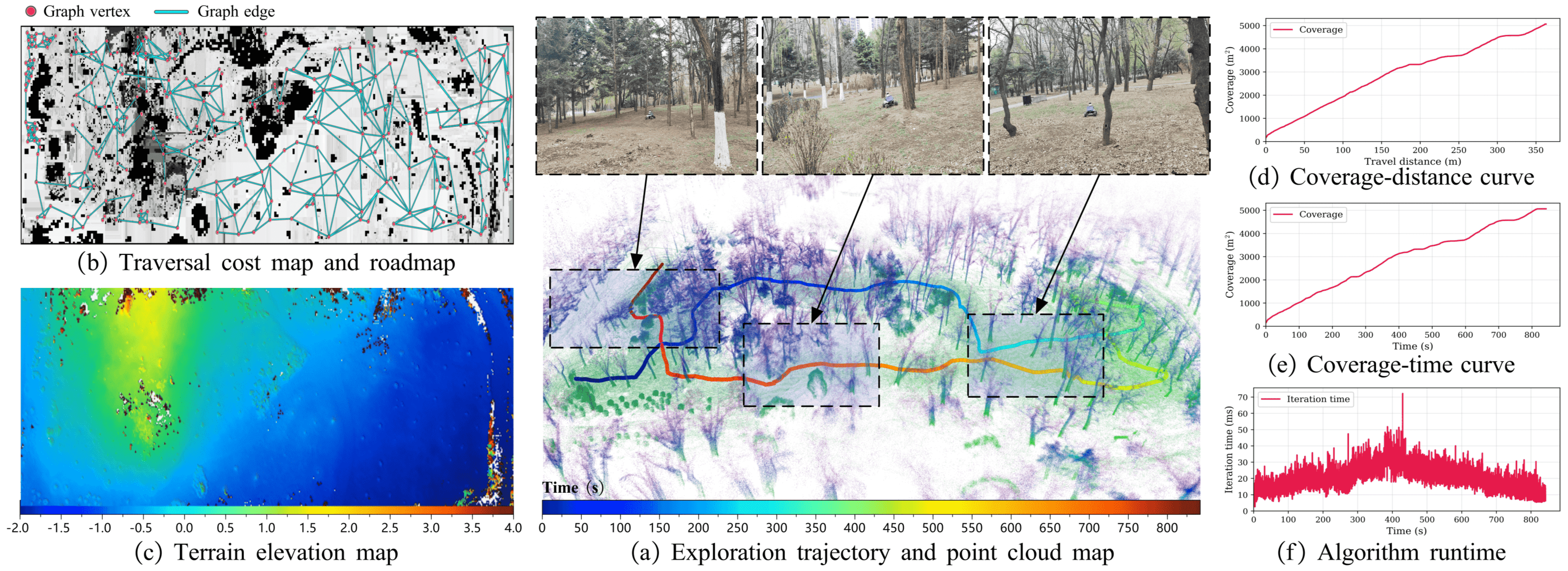}
	\caption{Medium-scale field exploration in uneven woodland. (a) Exploration trajectory. (b) Traversability cost map and roadmap after exploration. (c) Elevation map of the terrain. (d) Coverage versus movement distance. (e) Coverage versus runtime. (f) Algorithm iteration time versus runtime.}
	\label{fig:real_medium}
	\vspace{-0.4cm}
\end{figure*}
\begin{figure*}[t]
	\centering
	\includegraphics[width=0.96\textwidth]{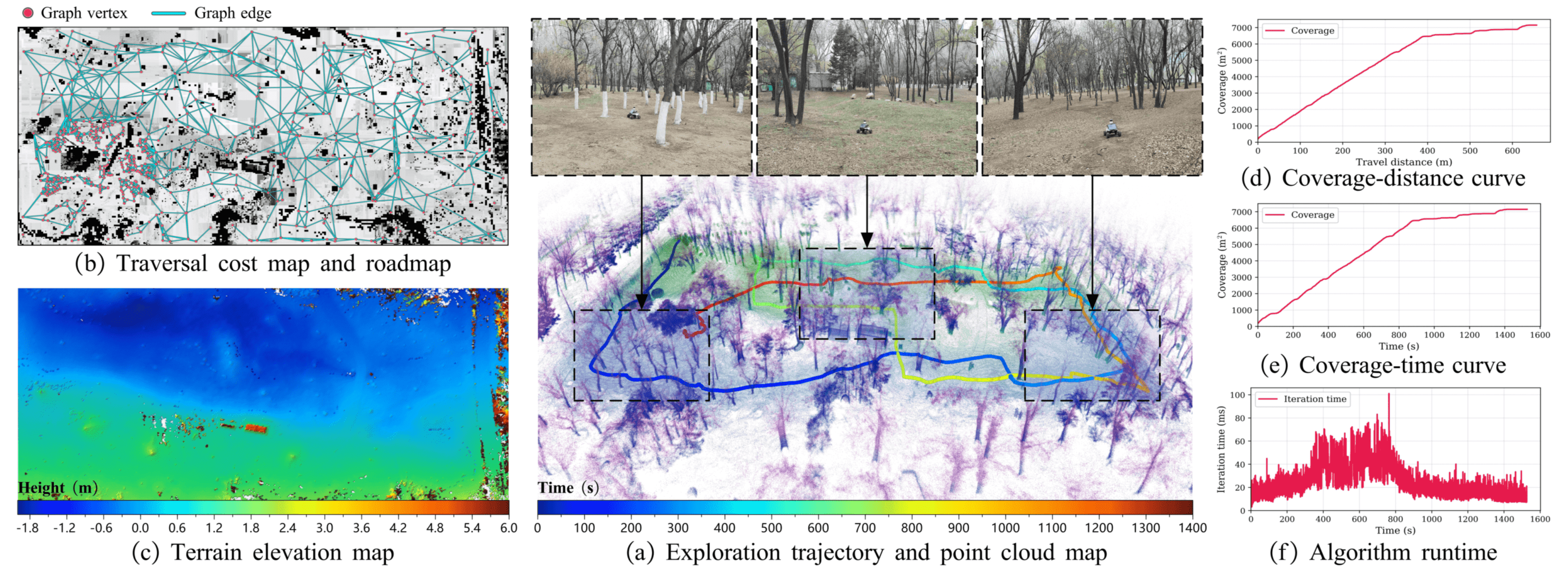}
	\caption{Large-scale field exploration in complex rugged terrain. (a) Exploration trajectory. (b) Traversability cost map and roadmap after exploration. (c) Elevation map of the terrain. (d) Coverage versus movement distance. (e) Coverage versus runtime. (f) Algorithm iteration time versus runtime.}
	\label{fig:real_large}
	\vspace{-0.4cm}
\end{figure*}

The three real-world scenes present different exploration difficulties. Scene 1 is a small open lawn of about \(50~\mathrm{m}\times50~\mathrm{m}\). It contains a few trees and mild local slopes, and it mainly evaluates fast expansion in weakly constrained open space. As shown in Fig.~\ref{fig:real_small}, the proposed method completes exploration in 365 s with a trajectory length of 156 m. The robot does not oscillate around local trees. It expands continuously along traversable areas. This result indicates that region-level unknown targets provide stable long-range guidance in open environments, and the dynamic roadmap connects remote targets with short paths.

Scene 2 is an uneven woodland of about \(125~\mathrm{m}\times50~\mathrm{m}\). Compared with Scene 1, it is much less structured. Trees, grass, rocks, and slopes coexist, and trunks form many narrow gaps. The maximum local height difference exceeds 2 m. This scene evaluates traversability reasoning under dense occlusion, terrain undulation, and complex obstacles. As shown in Fig.~\ref{fig:real_medium}, the proposed method completes exploration in 835 s with a trajectory length of 362 m. The robot continuously selects traversable passages in dense trees and avoids trunks, rocks, and high-cost slopes. This result shows that variable-voxel ground fitting can recover continuous ground from sparse LiDAR observations, while 8-bit obstacle encoding distinguishes vertical trunk obstacles from real slopes. The planner can therefore generate safe executable goals in complex woodland.

Scene 3 is the most challenging large-scale rugged terrain, with an area of about \(125~\mathrm{m}\times75~\mathrm{m}\). The terrain changes more frequently and severely, obstacle types are diverse, and LiDAR occlusion is strong. The maximum ground height difference exceeds 5 m. This scene evaluates the complete system after the map grows over a large area. As shown in Fig.~\ref{fig:real_large}, the proposed method completes exploration in 1524 s with a trajectory length of 757 m. Despite continuous slopes, occluded areas, and complex obstacle combinations, the robot keeps moving steadily and completes large-area coverage along traversable terrain. Sector segmentation organizes remote unknown space into clear visiting targets, and the near-dense, far-sparse roadmap maintains effective topological connectivity as the map expands.

Overall, the three real-world experiments validate different properties of the system. They demonstrate efficient expansion in open space, reliable terrain recognition in dense woodland, and online scalability in large rugged terrain. The average algorithm iteration times are 13.89 ms, 17.14 ms, and 20.71 ms in the three scenes, respectively. These values satisfy real-time requirements on the onboard edge-computing platform. Even in the largest scene, the system replans online while the numbers of roadmap vertices and unknown regions increase. The generated trajectories follow traversable areas, actively avoid dense trees, rocks, and high-cost slopes, and show stable growth in covered area over time and movement distance. These results indicate that the proposed method is effective not only in simulation benchmarks, but also in real outdoor environments with sensor noise, occlusion, and complex three-dimensional terrain.

\section{Conclusion}
\label{sec:conclusion}

This paper presented TASG-Explore, a terrain-aware autonomous exploration framework for ground robot operating in uneven outdoor environments. The key objective was to improve large-scale exploration efficiency while maintaining traversability-aware completeness and safety. To this end, TASG-Explore integrates hierarchical traversability analysis, incremental sector region segmentation, and dynamic topological roadmap planning. The proposed traversability analysis combines variable-voxel ground fitting with adaptive 8-bit obstacle encoding, enabling robust recovery of support surfaces and local obstacle structures from sparse and partially occluded observations. The incremental sector segmentation further characterizes the spatial distribution and exploration value of unknown regions, thereby providing efficient region-level guidance for planner. The dynamic roadmap with unknown topological hypotheses enables long-term topological map maintenance and unknown connectivity estimation, while the sector-guided planner leverages this structure to achieve efficient global exploration in large-scale scenes. Extensive benchmark experiments in caves, forests, and rugged hills demonstrate that TASG-Explore achieves superior overall exploration performance compared with existing baselines. Real-world experiments further validate its practicality for autonomous exploration in challenging outdoor uneven terrain. 
﻿

The proposed method is primarily designed for outdoor scenes that can be represented by a supporting ground. Therefore, it is not well suited to multi-layer environments with vertically overlapping traversable structures. Extending the environment representation and exploration planner to support multi-level scenes remains an important direction for future work.

\bibliographystyle{IEEEtran}
\bibliography{tro_references}

\end{document}